%% file: main.tex
\pdfoutput=1

\documentclass[11pt]{article}

\usepackage[]{ACL2023}

\usepackage{times}
\usepackage{latexsym}

\usepackage[T1]{fontenc}

\usepackage[utf8]{inputenc}

\usepackage{microtype}

\usepackage{inconsolata}

\usepackage{multirow}
\usepackage{varwidth}
\usepackage{makecell}
\usepackage{tabu, booktabs}
\usepackage{caption}
\usepackage{subcaption}
\usepackage{placeins}
\usepackage{enumitem}
\usepackage{inconsolata}
\usepackage{graphicx} 
\usepackage{array}
\usepackage{nicematrix,tikz}
\usepackage{lscape, adjustbox}
\usepackage{varwidth}
\usepackage{soul, colortbl}
\usepackage{stfloats}
\usepackage{float}
\usepackage{lipsum}
\usepackage{listings}
\usepackage{xcolor}
\usepackage{color, colortbl}
\usepackage[T1]{fontenc}
\usepackage{textcomp}
\usepackage{xspace}

\newcommand{\std}[1]{\ensuremath{\scriptsize\pm#1}}
\lstdefinestyle{mystyle}{
    backgroundcolor=\color{backcolour},   
    commentstyle=\color{codegreen},
    keywordstyle=\color{magenta},
    numberstyle=\tiny\color{codegray},
    stringstyle=\color{codepurple},
    basicstyle=\ttfamily\footnotesize,
    breakatwhitespace=false,         
    breaklines=true,                 
    captionpos=b,                    
    keepspaces=true,                 
    numbers=left,                    
    numbersep=5pt,                  
    showspaces=false,                
    showstringspaces=false,
    showtabs=false,                  
    tabsize=4
}
\definecolor{Gray}{gray}{0.88}
\usepackage[most]{tcolorbox}
\usepackage{listings}

\definecolor{mycolor}{RGB}{255, 230, 230} 

\lstdefinestyle{mystyle}{
  basicstyle=\ttfamily,
  columns=fullflexible,
  keepspaces=true,
  upquote=true,
}

\newcommand{\dataset}{\textsc{GraphTMI}\xspace}

\title{Which Modality should I use - Text, Motif, or Image? : Understanding Graphs with Large Language Models}

\author{Debarati Das \quad Ishaan Gupta \quad Jaideep Srivastava \quad Dongyeop Kang \\
        Department of Computer Science, University of Minnesota \\
               \texttt{\{das00015, gupta737, srivasta, dongyeop}\}@umn.edu\\}

\begin{document}
\maketitle
\begin{abstract}
Our research integrates graph data with Large Language Models (LLMs), which, despite their advancements in various fields using large text corpora, face limitations in encoding entire graphs due to context size constraints. This paper introduces a new approach to encoding a graph with diverse modalities, such as text, image, and motif, coupled with prompts to approximate a graph's global connectivity, thereby enhancing LLMs' efficiency in processing complex graph structures. The study also presents \dataset, a novel benchmark for evaluating LLMs in graph structure analysis, focusing on homophily, motif presence, and graph difficulty. Key findings indicate that the image modality, especially with vision-language models like GPT-4V, is superior to text in balancing token limits and preserving essential information and outperforms prior graph neural net (GNN) encoders. Furthermore, the research assesses how various factors affect the performance of each encoding modality and outlines the existing challenges and potential future developments for LLMs in graph understanding and reasoning tasks.
All data will be publicly available upon acceptance.
\end{abstract}
\section{Introduction}
\input{tex/intro}
\section{Setups}
\input{tex/data}

\section{Proposed Encoders with Different Modalities}
\input{tex/proposed_methods}

\input{tex/exp_setup}

\section{Results}

\input{tex/results_section}

\section{Related Work}
\input{tex/related_work}

\section{Conclusion and Future Work}
\input{tex/conclusion}

\input{tex/futurework}

\section*{Limitations}
\input{tex/limitations}

\bibliography{custom}
\bibliographystyle{acl_natbib}
\appendix
\input{tex/appendix}
\end{document}

%% file: tex/intro.tex

Large Language Models (LLMs) are increasingly utilized in areas with inherent graph structures like social network analysis \cite{mislove2007measurement}, drug discovery \cite{vishveshwara2002protein}, and recommendation systems \cite{melville2010recommender}, but they face limitations due to their reliance on unstructured text and challenges in incorporating new data post-training \cite{zhang2023siren,lewis2020retrieval,pan2023unifying}. Graph-structured data can address these issues, providing a nuanced and flexible representation of real-world relationships.

\begin{figure}[t]
  \centering
\includegraphics[width = \linewidth,trim={7.2cm 1.4cm 9.9cm 2.6cm},clip]{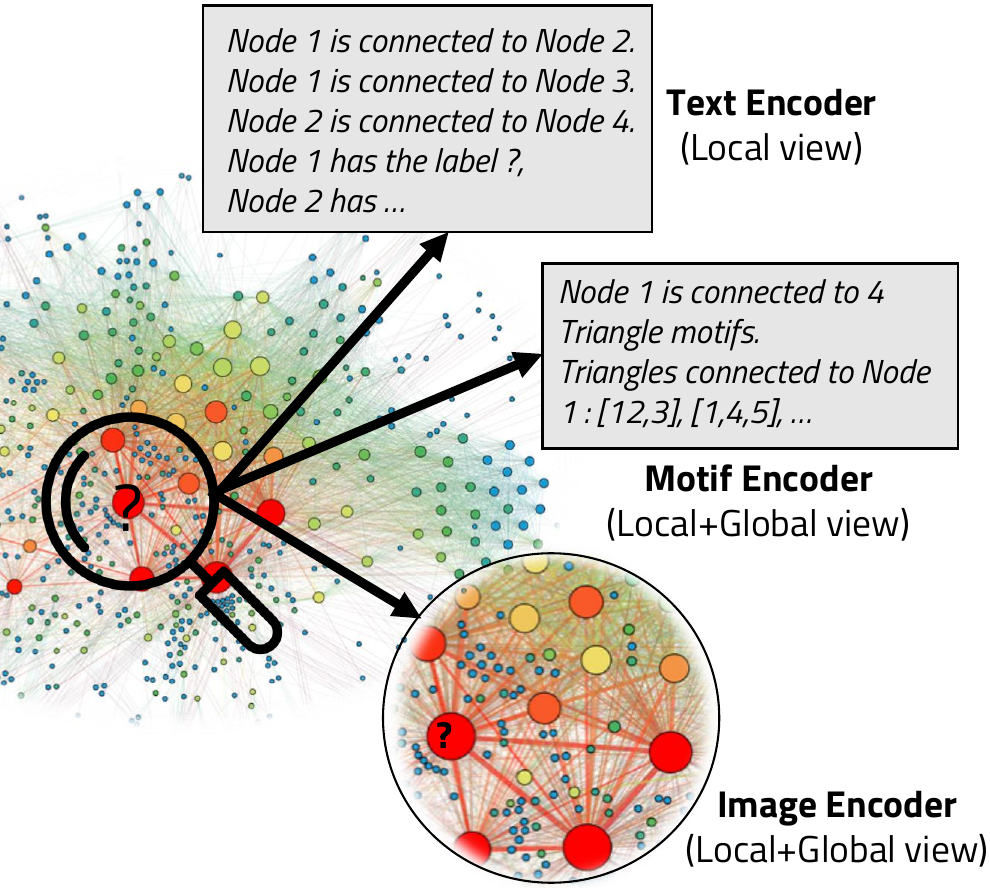}
  \caption{Input modality encoding for graphs impacts node classification, with text modality offering detailed information from a \textit{local} point of view but violating the input context limitations for LLMs due to verbosity. Motif modality provides \textit{local and global context}, while image modality gives a comprehensive \textit{global} view, efficiently processed by GPT-4V, which integrates capabilities from both vision and text.}
  \label{fig:fig1}
\end{figure}

While there has been progress in interpreting multi-modal information \cite{yin2023survey}, integrating graph understanding into LLMs remains a developing area. Current challenges include LLMs' difficulty directly processing complex graph-structured data, necessitating innovative input encoding and prompt design \cite{fatemi2023talk,chen2023exploring} for various graph tasks. Current research typically employs limited setups with small real-world graphs \cite{guo2023gpt4graph} or synthetic ones \cite{wang2023can}, exposing a gap in effectively incorporating large real-world graphs into LLMs, owing to their context window constraints. This suggests that text-only encoding may not be the optimal approach for complex, large graph structures.

This paper investigates the impact of different modalities for encoding global and local graph structures, focusing on node classification tasks, and compares three modalities: \textit{Text}, \textit{Motif}, and \textit{Image} (See Figure \ref{fig:fig1}). Text modality offers detailed local insights but becomes verbose for large graphs \cite{bubeck2023sparks}, often exceeding the input limits of models like GPT-4. The Motif modality is suggested to address this, capturing essential patterns in a node's vicinity for a balanced local-global perspective. Additionally, Image modality is proposed, utilizing fewer tokens to convey a more global view of the node's neighborhood, a method enhanced by the vision capabilities of the newly released GPT-4V \cite{openai_gpt4}. 
Finding the optimal prompt input format is a notably complex challenge, with text modality encoding requiring extensive exploration compared to the simpler, more human-readable image modality. In our evaluations, we balance informativeness and prompt conciseness across all modalities using a combination of metrics. 


Our \textbf{main contributions} are as follows:
\begin{itemize}[noitemsep,topsep=0pt,leftmargin=*]
    \item We conduct breadth-first analysis of various modalities, such as text, image, and motif, in graph-structure prompting, utilizing large language and vision-language models for node classification tasks.
    \item We also perform a depth-first analysis of how different factors influence the performance of each encoding modality.
    \item We introduce \dataset, a novel graph benchmark featuring a hierarchy of graphs, associated prompts, and encoding modalities designed to further the community's understanding of graph structure effects using LLMs. 
\end{itemize}

\textbf{Some key findings}: 1) When balancing the constraint of token limits while preserving crucial information, the image modality is more effective than the text modality for graph-related tasks. 2) The choice of encoding modality for graph task classification depends on the task's difficulty, assessed by homophily and motif counts, with image modality being optimal for medium-difficulty tasks and motif modality for harder ones. 3) Factors like edge encoding function, graph structure, and graph sampling techniques impact the performance of node classification using text modality. 4) Motif attachment information has a more significant impact on node classification than motif count information. 5) Image representation correlated with human readability positively impacts node classification performance.  

Our research shows that while LLMs are progressing in graph data processing, they still fall short compared to Graph Neural Networks in managing real-world graphs, emphasizing both the current limitations and the future potential of LLMs with image understanding in graph interpretation and reasoning tasks.


%% file: tex/data.tex
\subsection{Seed Datasets}
\label{sec:datasets}
\begin{table}[t] 
\centering
\setlength{\tabcolsep}{4pt} 
\begin{tabular}{lccc}
\toprule
\textbf{Properties} & \textbf{CORA} & \textbf{Citeseer} & \textbf{Pubmed} \\
\midrule
Classes & 7 & 6 & 3 \\
Nodes & 2,708 & 3,327 & 19,717 \\
Edges & 5,278 & 4,552 & 44,324 \\
Density & 0.0014 & 0.0008 & 0.0002 \\
Avg deg & 3.89 & 2.74 & 4.49 \\
Clust coeff & 0.24 & 0.14 & 0.06 \\
Diameter & $\infty$ & $\infty$ & 18 \\
Components & 78 & 438 & 1 \\
2-hop nodes & 36 & 15 & 60 \\
\bottomrule
\end{tabular}
\caption{Comparison of network properties of popular citation network datasets CORA, Citeseer and Pubmed.} 
\label{tab:dataset_comparison}
\end{table}

We experiment with three citation network datasets, which are popular node classification benchmarks, \textsc{CORA} \cite{mccallum2000automating} with seven categories : [0-Rule Learning, 1-Neural Networks, 2-Case-Based, 3-Genetic Algorithms, 4-Theory, 5-Reinforcement Learning, and 6-Probabilistic Methods], \textsc{CITESEER} \cite{giles1998citeseer} with six categories of areas in Computer Science: [0-Agents, 1-ML, 2-IR, 3-DB, 4-HCI, 5-AI] and \textsc{PUBMED} \cite{sen2008collective} that consists of scientific journals collected from the PubMed database with the following three categories: [0-Diabetes Mellitus, Experimental, 1-Diabetes Mellitus Type 1, 2-Diabetes Mellitus Type 2]. This paper focuses solely on the structural information of graphs for node classification. Hence, our experiments exclusively utilize node and label IDs.
\subsection{Evaluation Metrics}
This paper assesses the performance of node classification using four metrics chosen to balance the tradeoff between the encoding's informativeness and verbosity. The metrics used are Accuracy rate (which should increase $\uparrow$), Denial rate (which should decrease $\downarrow$), Mismatch rate (which should decrease $\downarrow$), indicating the prompt's informativeness, and Token limit fraction (which should decrease $\downarrow$), reflecting the prompt's verbosity. 

\noindent \textbf{Accuracy Rate \(A\)}: This metric indicates the LLM's performance on the task of node classification.
\begin{equation}
A = \frac{\text{No. of correct predictions}}{\text{Total no. of samples}}
\end{equation}

\noindent \textbf{Mismatch Rate \(M\)}: This metric indicates the degree of misclassification by LLM (when the ground truth value is not the same as the predicted value).
\begin{equation}
M = \frac{\text{No. of incorrect predictions}}{\text{Total no. of samples}}
\end{equation}

\begin{figure*}
  \centering
\includegraphics[width = 0.9\linewidth,trim={0cm 0cm 6.7cm 0cm},clip]{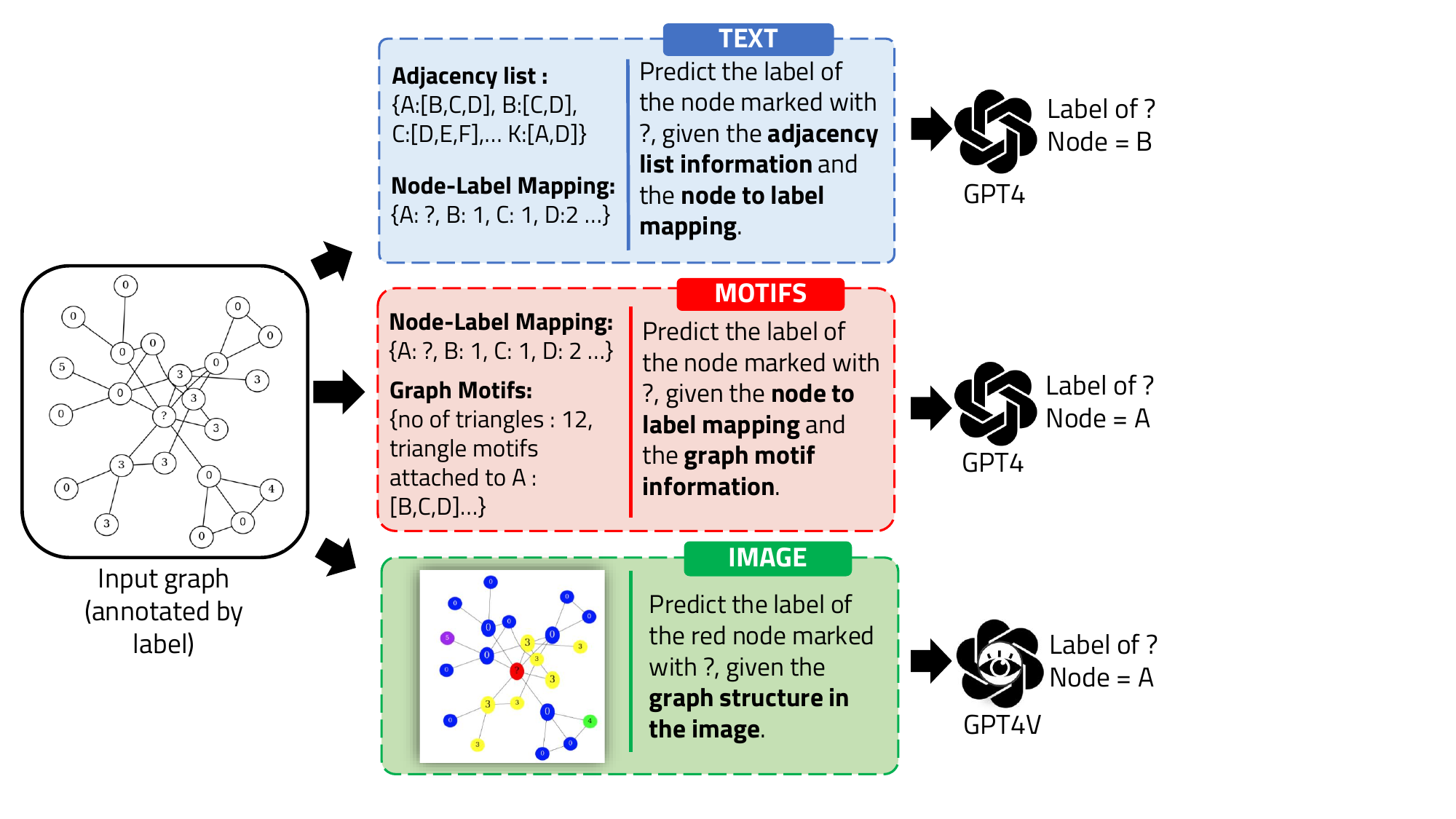}
\vspace{-5mm}
  \caption{Node Classification on a Graph using different input modality encodings like Text, Motif, and Image. }
  \label{fig:fig2}
\end{figure*}

\noindent \textbf{Denial Rate \(D\)}: When we craft our prompt, we instruct the LLM to return -1 if it cannot predict the label of the \(?\) node (node to be classified). The denial rate metric describes the rate of failure of the LLM (when the predicted value is -1).
\begin{equation}
D = \frac{\text{No. of predictions = -1}}{\text{Total no. of samples}}
\end{equation}
\begin{equation}
 1-A = M + D
\end{equation}

\noindent \textbf{Token Limit Fraction $T$}: This metric evaluates how effectively a Large Language Model's encoding modality uses its input context window, specifically focusing on the constraints imposed by the fixed-size attention window in transformer-based models like GPT-4 and GPT-4V. These constraints, dictated by the model's neural network architecture, limit the number of tokens that can be processed simultaneously, impacting both computational cost and performance.
\begin{equation}
T = \frac{\text{Number of usage tokens}}{\text{Token limit constraint for the model}}
\end{equation} 
\subsection{Graph Encoder Baselines} We compare our LLM models, which use different encoding modalities, to traditional graph learning models \textsc{GCN}\cite{kipf2016semi}, \textsc{GraphSage}\cite{hamilton2017inductive} and \textsc{GAT} \cite{velivckovic2017graph} as a baseline for node classification tasks. We provide the training details for the GNN models in Appendix \ref{sec:B}.

%% file: tex/proposed_methods.tex
\begin{figure*}[t]
     \centering
     \begin{subfigure}[b]{0.32\textwidth}
         \centering
\includegraphics[width=0.8\columnwidth]{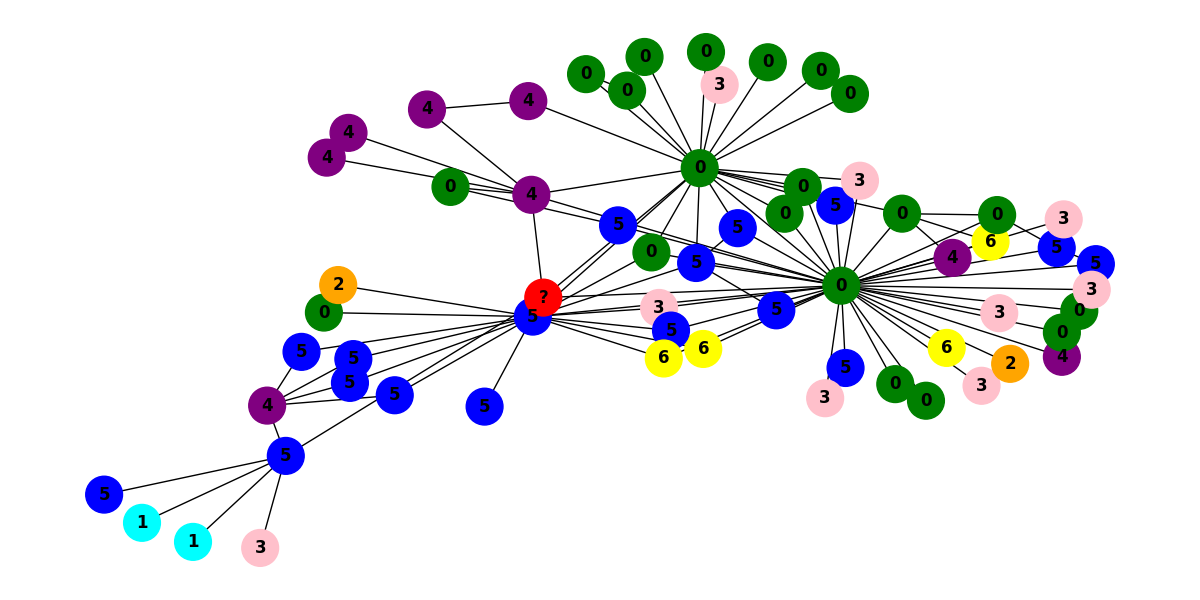}         \caption{\textbf{Original NetworkX Graph} \cite{hagberg2008exploring} - no image setting changes made}
     \end{subfigure}
     \hfill
     \begin{subfigure}[b]{0.32\textwidth}
         \centering
\includegraphics[width=0.8\columnwidth]{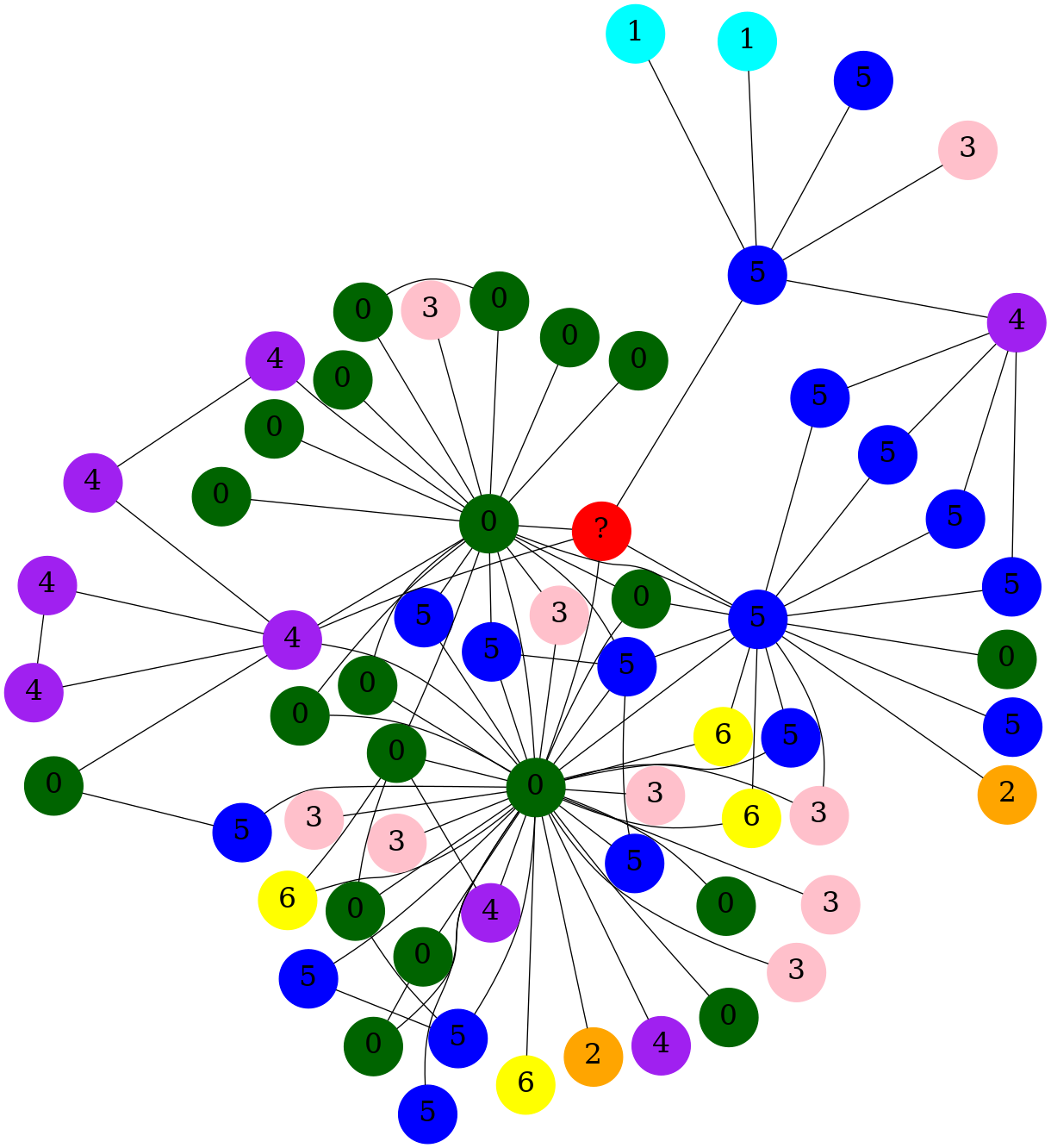}         \caption{\textbf{Node Size Increase} - changing the rendering using GraphViz \cite{ellson2002graphviz} and increasing the size of each node in the graph, increasing node and label clarity}
     \end{subfigure}
     \hfill
     \begin{subfigure}[b]{0.32\textwidth}
         \centering
         \includegraphics[width=0.8\columnwidth]{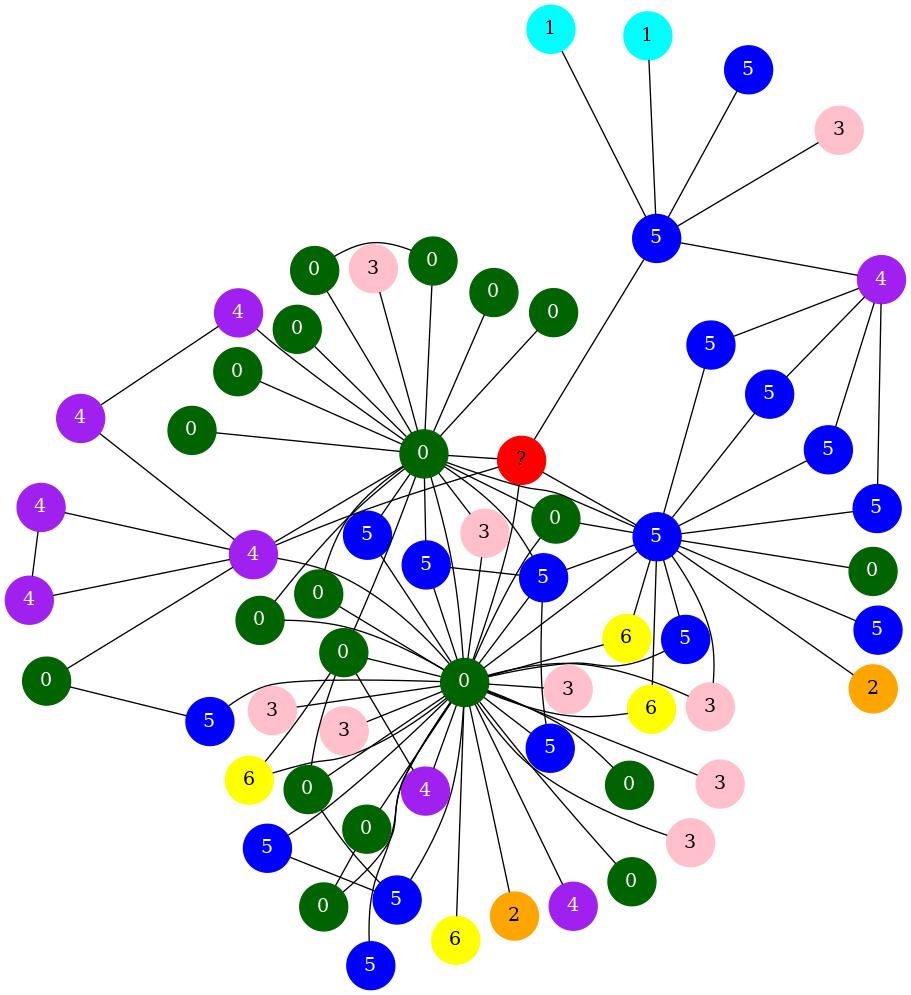}
         \caption{\textbf{Contrasting Text Color} - contrasting the textual labels with the node color will make the labels more ``human-readable'' }
     \end{subfigure}
\hfill
     \begin{subfigure}[b]{0.32\textwidth}
         \centering
         \includegraphics[width=0.8\columnwidth]{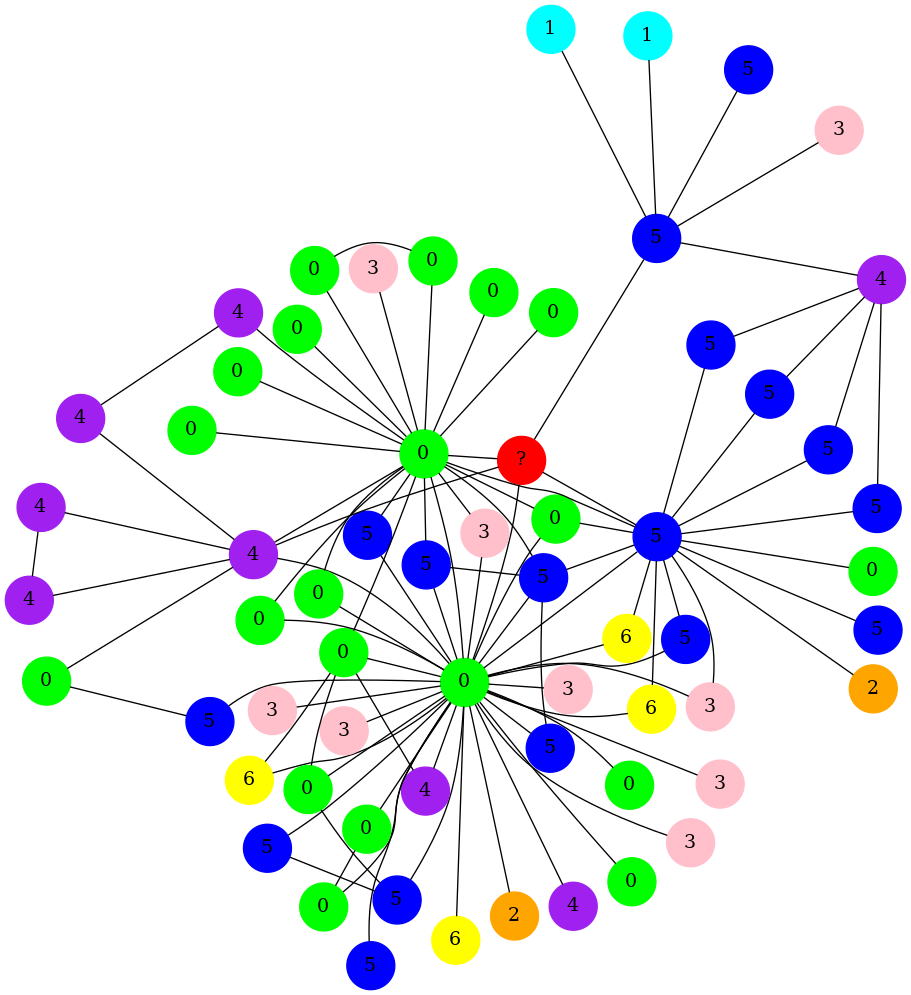}
         \caption{\textbf{Distinctive Node Colors} - choosing distinct colors allows the VLM to distinguish differently labeled nodes better}
     \end{subfigure}
\hfill
     \begin{subfigure}[b]{0.32\textwidth}
         \centering
         \includegraphics[width=0.8\columnwidth]{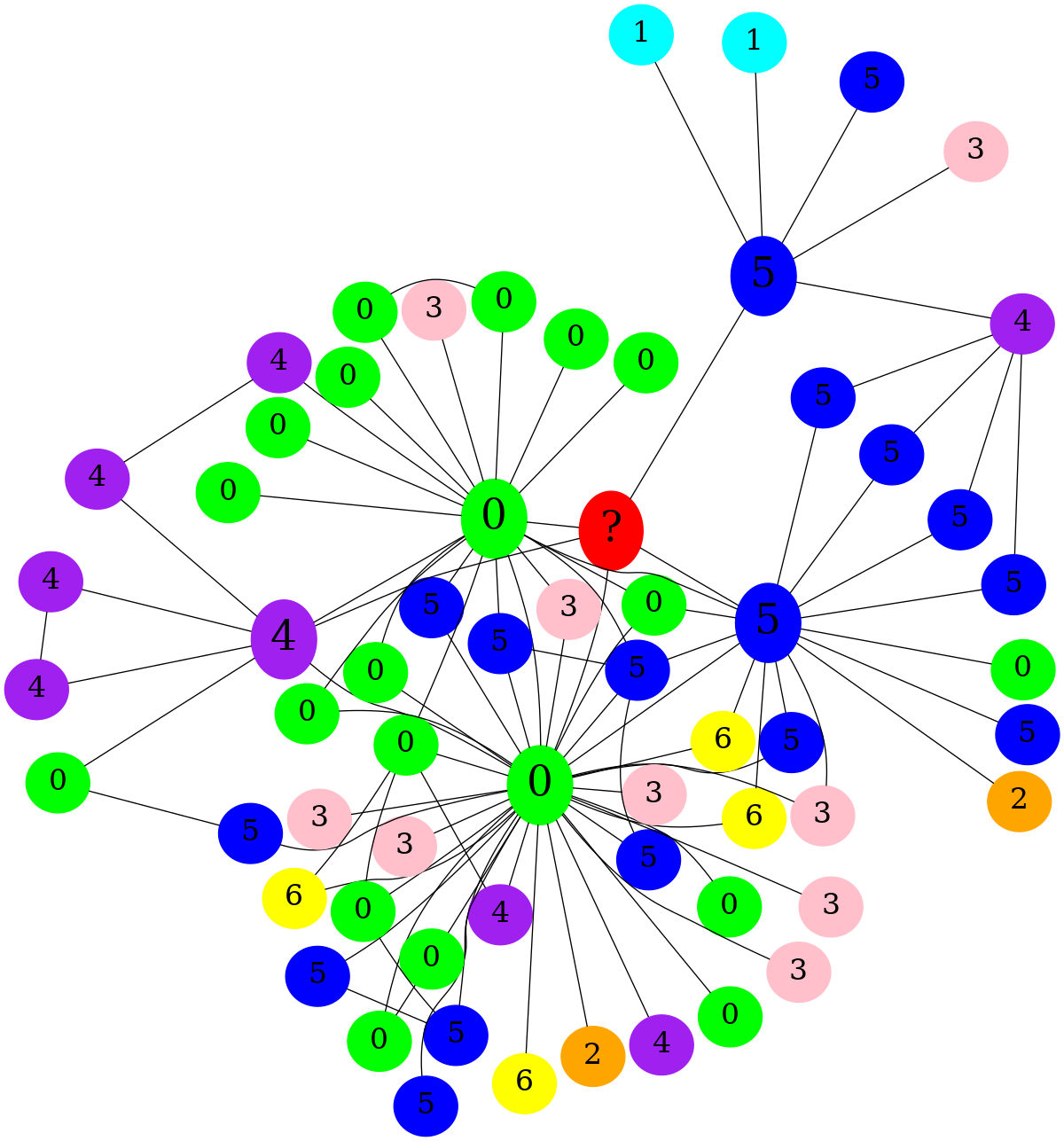}
         \caption{\textbf{Node Size increase based on 1-Hop-distance} - increasing the size of nodes near the vicinity of the un-labeled node might allow the VLM to prioritize the data of nodes of greater size as opposed to nodes further away}
     \end{subfigure}
\hfill
     \begin{subfigure}[b]{0.32\textwidth}
         \centering
         \includegraphics[width=0.8\columnwidth]{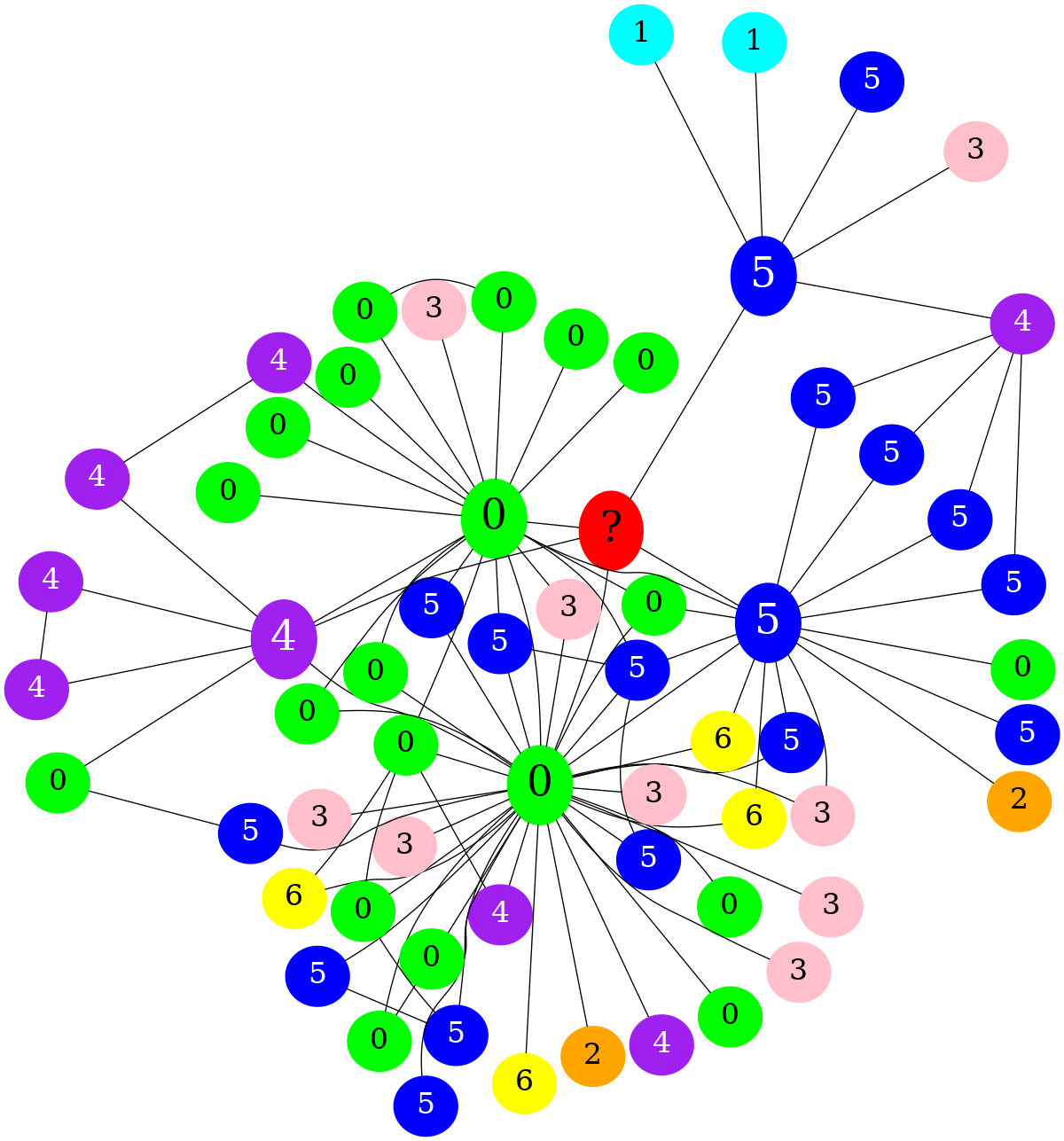}
         \caption{\textbf{Aggregate all changes} - all changes from b) to e) are applied to maximize image clarity  }
     \end{subfigure}
\caption{Image representation changes were applied sequentially on a graph, and we observed a distinct \textit{increase from (a) to (f) in human readability and understanding of the graph} structure.}

\label{tab:img_modality_changes}
\end{figure*}
Graph encoding is crucial for converting graph-structured data into a sequence format that language models can process. As shown in Figure \ref{fig:fig2}, the experimental setup involves using a modality encoder to input the graph structure and a graph query, such as predicting a node's label in node classification tasks. The graph structure is encoded according to the chosen modality (text and motif using GPT-4 and image using GPT-4V) and then passed as a prompt to the LLMs to generate the required label.

\subsection {Text Encoder}
In encoding graphs as text, nodes are mapped to labels using a dictionary format, and different edge-encoding representations \cite{guo2023gpt4graph,fatemi2023talk} are experimented with (Table \ref{tab:edge_representations}), providing \textit{local context} through edge connections and node labels to GPT-4 \cite{openai_gpt4}. However, larger graphs can lead to verbose text encodings, which may exceed LLM input limits. We evaluate the impact of graph structure on classification \cite{yasir2023examining, palowitch2022graphworld} by analyzing real-world citation datasets like \textsc{Pubmed}, \textsc{Citeseer}, and \textsc{CORA}, each with distinct network properties (definitions for these are provided in Table \ref{tab:graph_properties} and distinguished through Table \ref{tab:dataset_comparison}). \textsc{Pubmed} is the largest and most connected but has the lowest clustering coefficient, indicating less local clustering. In contrast, \textsc{Citeseer} is highly fragmented with many disconnected components, while \textsc{CORA}, the smallest network, exhibits the highest density and clustering coefficient, suggesting strong local connectivity. 
Additionally, the research examines graph sampling techniques like ego graph \cite{stolz2021predicting} and forest fire sampling \cite{leskovec2006sampling}, crucial due to LLMs' limited context window and complex real-world graphs \cite{wei2022evaluating}. These methods vary in their effectiveness, with Forest Fire sampling providing a broad network view, suitable for large networks like \textsc{Pubmed}, and Ego graph sampling excelling in revealing local community structures in more clustered and locally dense networks like \textsc{CORA} and \textsc{Citeseer}.

\begin{figure*}
     \centering
     \small
     \begin{subfigure}[b]{0.31\textwidth}
         \centering
\includegraphics[width=0.8\columnwidth]{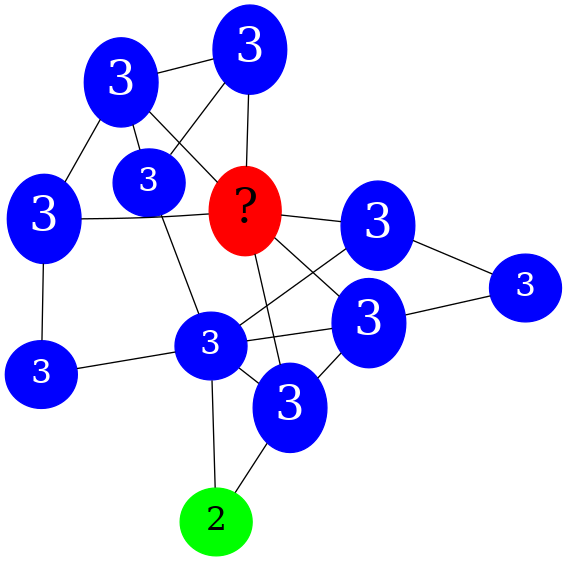}
\caption{\textbf{EASY} - ($\#$distinct labels $<3$) and ($\#$motifs $\leq 10$)}
     \end{subfigure}
     \hfill
     \begin{subfigure}[b]{0.31\textwidth}
         \centering
\includegraphics[width=0.8\columnwidth]{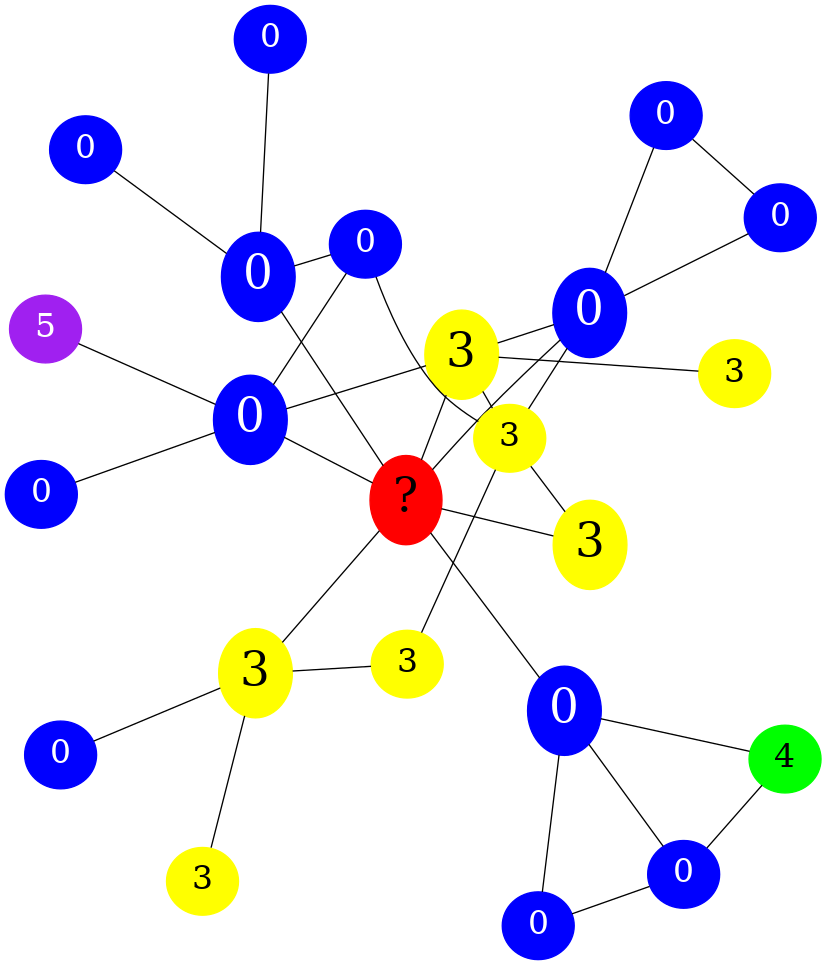}         
\caption{\textbf{MEDIUM} - ($3\leq \#$ distinct labels $<5$) and ($10 < \#$ motifs $\leq 20$)}
     \end{subfigure}
     \hfill
     \begin{subfigure}[b]{0.31\textwidth}
         \centering
         \includegraphics[width=0.8\columnwidth]{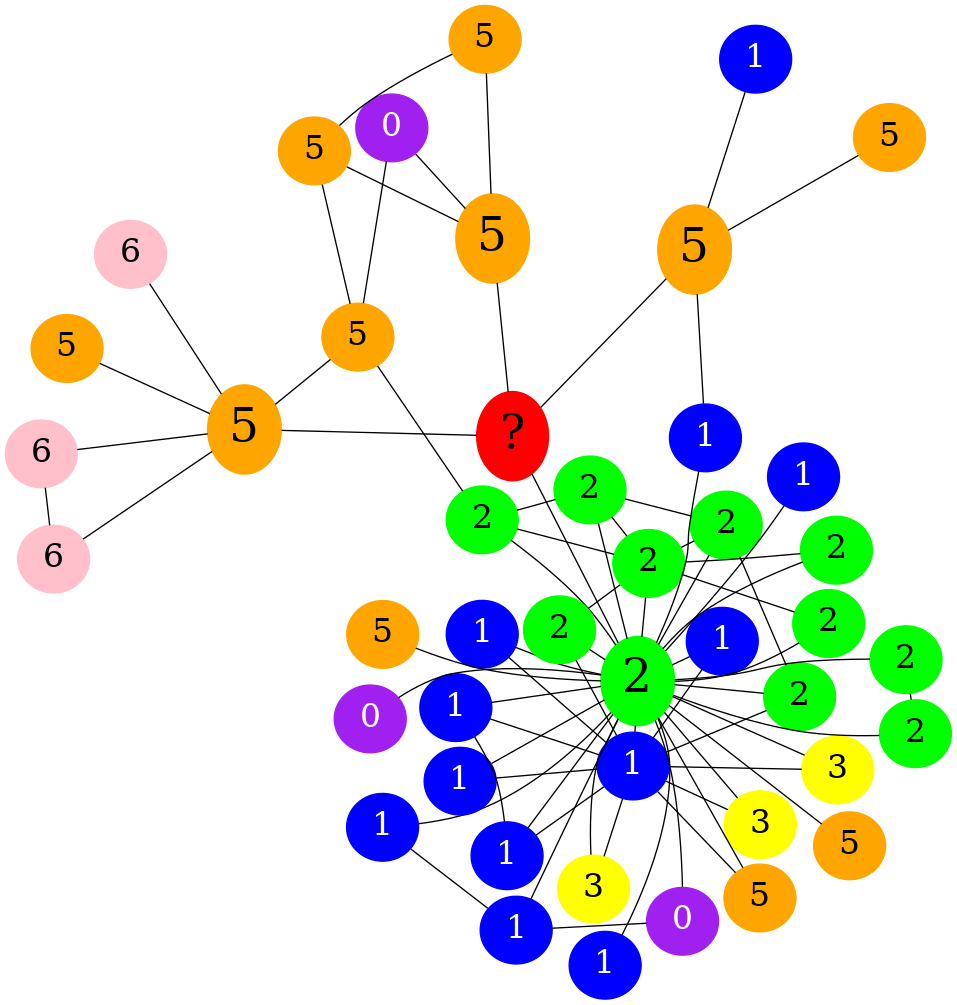}
         \caption{HARD - ($\#$distinct labels $\geq 5$) and ($\#$ motifs $\geq 20$)}
     \end{subfigure}
\caption{Classifying graph task difficulty based on the criteria of Homophily and Number of Motifs yields a dataset of EASY, MEDIUM, and HARD graph problems and their associated modality encodings and classifications. This benchmark is called the \textsc{GraphTMI} dataset. }
\label{tab:task_hierarchy}
\end{figure*}

\subsection {Motif Encoder}
Network motifs, recurring patterns in social and biological networks \cite{milo2002network,carrington2005models,holland1974statistical}, are pivotal in understanding local structures and behaviors. In LLMs, motif modality encoding leverages these motifs to provide \textit{local and global context}, aiding in classifying unlabeled nodes \cite{yangnode}. This process entails mapping nodes to labels using a dictionary format and identifying key motifs around the unlabeled node, which are inputted into GPT-4 as graph-motif information (detailed in Table \ref{tab:motif_representations}). Differentiating between the count and specific members of motifs like stars, triangles, and cliques in a graph, our approach posits that a node's connection to influential motifs, such as being central in a star for network influence or part of a triangle or clique for close community ties, can significantly affect its classification by revealing key aspects of the network structure.
\subsection {Image Encoder}

Adopting the idea that ``a picture is worth a thousand words'', the image modality in graph analysis uses visual representations to outperform text in depicting structures, networks, labels, and spatial relationships using fewer tokens. Vision-language models like GPT-4V \cite{openai_gpt4v} interpret these graph images, offering a \textit{global context} of the graph's structure. GPT-4V, a multimodal model, merges visual interpretation with language processing, underscoring the importance of image representation in enhancing node classification. Our experiments involved using graph rendering methods to generate images with color-coded nodes, with a focus on improving human readability through various image modifications (Figure \ref{tab:img_modality_changes}). These changes were evaluated for their impact on node classification, highlighting the critical role of visual representation in this modality.

\begin{table*}[ht]
\centering
\begin{tabular}{p{1.5 cm}|p{1.5cm}|ccc}
\toprule
& \textbf{Model} & \textbf{Cora} & \textbf{Citeseer} & \textbf{Pubmed} \\
\midrule
\multirow{3}{1.5cm}{GNN Baselines} & GCN & 0.7584 \std{0.121} & 0.6102 \std{0.087} & 0.7546 \std{0.076} \\
& GAT & \underline{0.7989} \std{0.092} & 0.6583 \std{0.074} & 0.7490 \std{0.060} \\
& GraphSage & 0.7719 \std{0.124} & 0.6017 \std{0.103} & 0.7193 \std{0.076} \\
\midrule
\multirow{3}{1.5cm}{LLMs + Encoding Modality} & Text & \textbf{0.81} \std{0.04} \small [0.07 \std{0.03}]  & \textbf{0.75} \std{0.05} \small [0.07 \std{0.01}] & \textbf{0.83} \std{0.01} \small [0.08 \std{0.01}]\textsuperscript{*}\\

& Motif & 0.73 \std{0.06} \small [0.06 \std{0.01}] & 0.59 \std{0.01} \small [0.32 \std{0.02}] & 0.77 \std{0.006} \small [0.13 \std{0.04}]\\

& Image & 0.77 \std{0.05} \small [0.04 \std{0.02}]\textsuperscript{*} & \underline{0.71} \std{0.09} \small [0.06 \std{0.0}]\textsuperscript{*} & \underline{0.79} \std{0.03} \small [0.19 \std{0.01}] \\

\bottomrule
\end{tabular}\caption{ We report test accuracy rates of node classification across different datasets and denial rates $D$ in [brackets] for LLM models. \textsuperscript{*} indicates the lowest denial rate for each modality. The highest accuracy rate for the dataset is in bold, while the second highest is underlined. \textit{The text modality in LLMS is comparable to GNN baselines with image modality not far behind.}} 
\label{tab:accuracy_tab}
\end{table*}

%% file: tex/exp_setup.tex
\section{GraphTMI Benchmark Creation}
Our study reveals that the ease of node classification in graphs varies across different modalities, depending on the graph's ``difficulty,'' determined by motif count and homophily. Homophily \cite{mcpherson2001birds}, based on network theory, suggests that nodes are more likely to connect with similar nodes; thus, graphs with higher homophily (more nodes sharing the same label) are simpler to classify than those with more heterophily (diverse labels). This is illustrated through \textsc{CORA} dataset examples in Figure \ref{tab:task_hierarchy}. Graphs are categorized as ``easy,'' ``medium,'' or ``hard'' based on the diversity of labels. Additionally, graphs with more network motifs are considered more complex and challenging for classification \cite{tu2018network}. The "task difficulty" is defined across eight categories ($2^3 = 8$), with the final difficulty level determined by the higher of the two criteria, homophily or motif count. This led to the creation of \textsc{GraphTMI}, a new benchmark dataset that includes various graph structures along with their respective modalities (text, motif, and image), prompts, and LLM classifications, thereby providing deeper insights into how different graphs affect LLM prompting. Specific statistics are given in Appendix \ref{sec:A}.


%% file: tex/results_section.tex
\subsection{Results Across All Modalities}
\noindent \textbf{Comparing Node Classification Accuracies between Graph baselines and LLM models :} Table \ref{tab:accuracy_tab} compares node classification accuracies of traditional GNN methods and LLM baselines across datasets, assessing if LLMs match conventional techniques. Limited by GPT-V's rate limit, the study used 50 ego graphs, with more extensive results in Appendix \ref{sec:A}. The \textit{text modality of LLMs performs comparably to graph baselines in all datasets, with the image modality close behind, indicating LLMs' potential in graph analysis}. In larger datasets like \textsc{Pubmed}, the image modality showed a higher denial rate, possibly due to overcrowding in larger subgraphs, leading to more frequent classification denials by the LLM.

\begin{figure*}[ht!]
    \centering
    \begin{minipage}{0.45\textwidth} 
        \centering
        \includegraphics[width=0.9\linewidth]{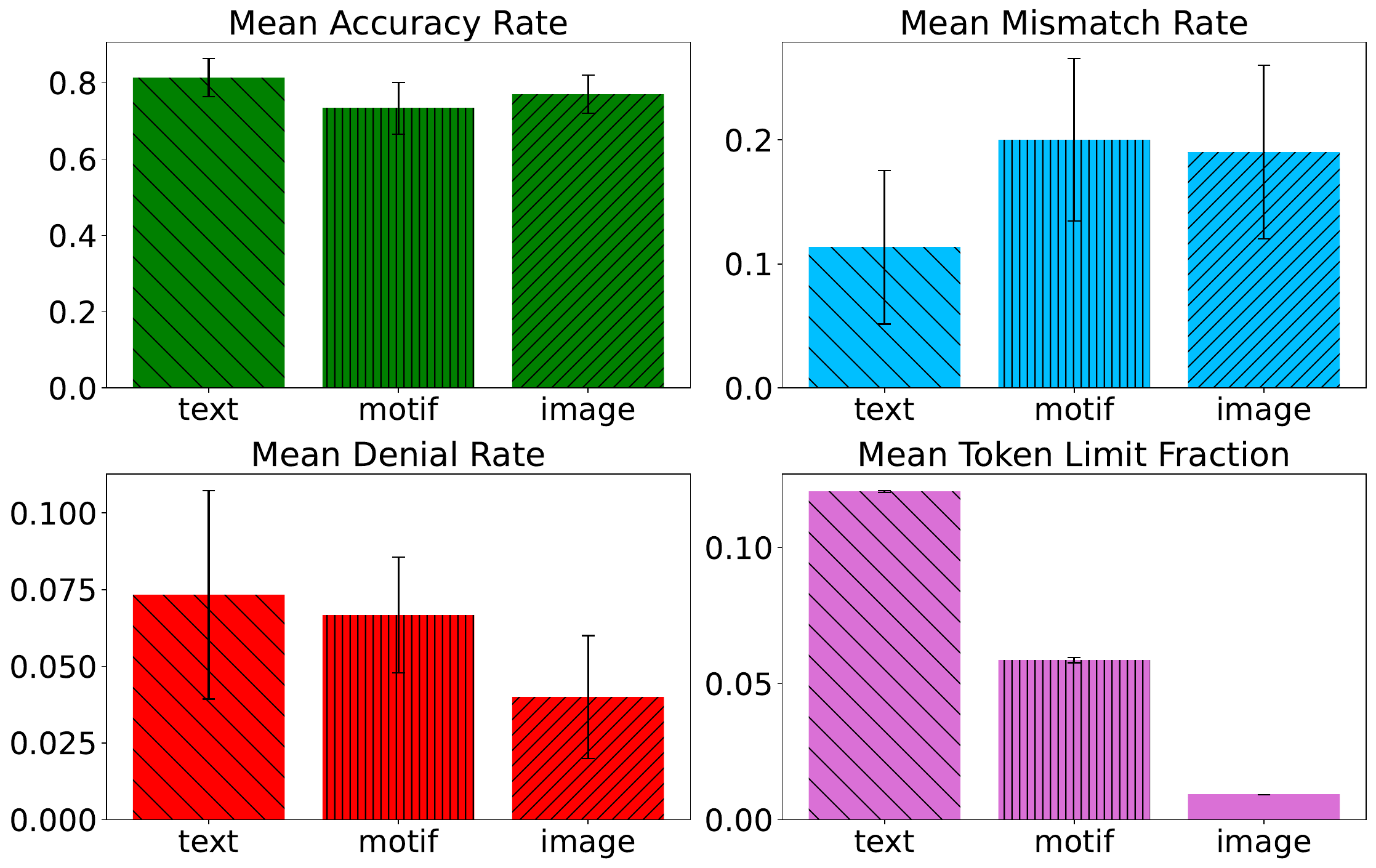}
        \caption{We observe that while the \textit{text and image modalities have similar accuracy rates, the motif modality exhibits the highest mismatch rate, and the image modality stands out with the lowest denial rate and token limit fraction}, as depicted along the mean metrics (y-axis) against each modality type (x-axis)}
        \label{fig:fig8}
    \end{minipage}
    \hspace{0.04\textwidth} 
    \begin{minipage}{0.45\textwidth} 
        \centering
        \includegraphics[width=0.7\linewidth]{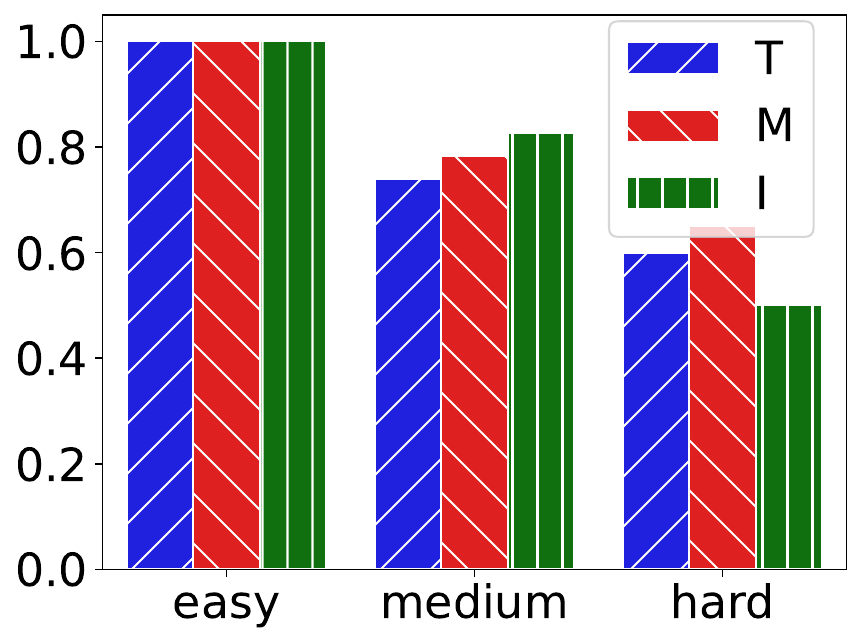}
        \caption{Modality encoder trends (T= Text, M= Motif, I= Image) with graph task difficulty based on homophily and no. of motifs, highlight the \textit{significance of integrating local and global information in LLM processing.}}
        \label{fig:fig9}
    \end{minipage}
\end{figure*}

\noindent \textbf{Comparing Node Classification Performance across Encoding Modalities}: Figure \ref{fig:fig8} compares node classification across encoding modalities, focusing on accuracy, mismatch, denial rates, and token limit fraction. The text modality shows high accuracy but struggles with a high denial rate and token limit fraction, likely due to verbose inputs that confuse the LLM. In contrast, the image modality offers similar accuracy but with lower denial rates and token limit fractions, indicating the \textit{image modality's effectiveness in providing a concise, global context that the LLM processes more efficiently}. This can be further emphasized through our qualitative analysis in Table \ref{tab:qual}.
\begin{figure*}
     \centering
     \begin{subfigure}[b]{0.25\textwidth}
         \centering
    \includegraphics[width=\columnwidth]{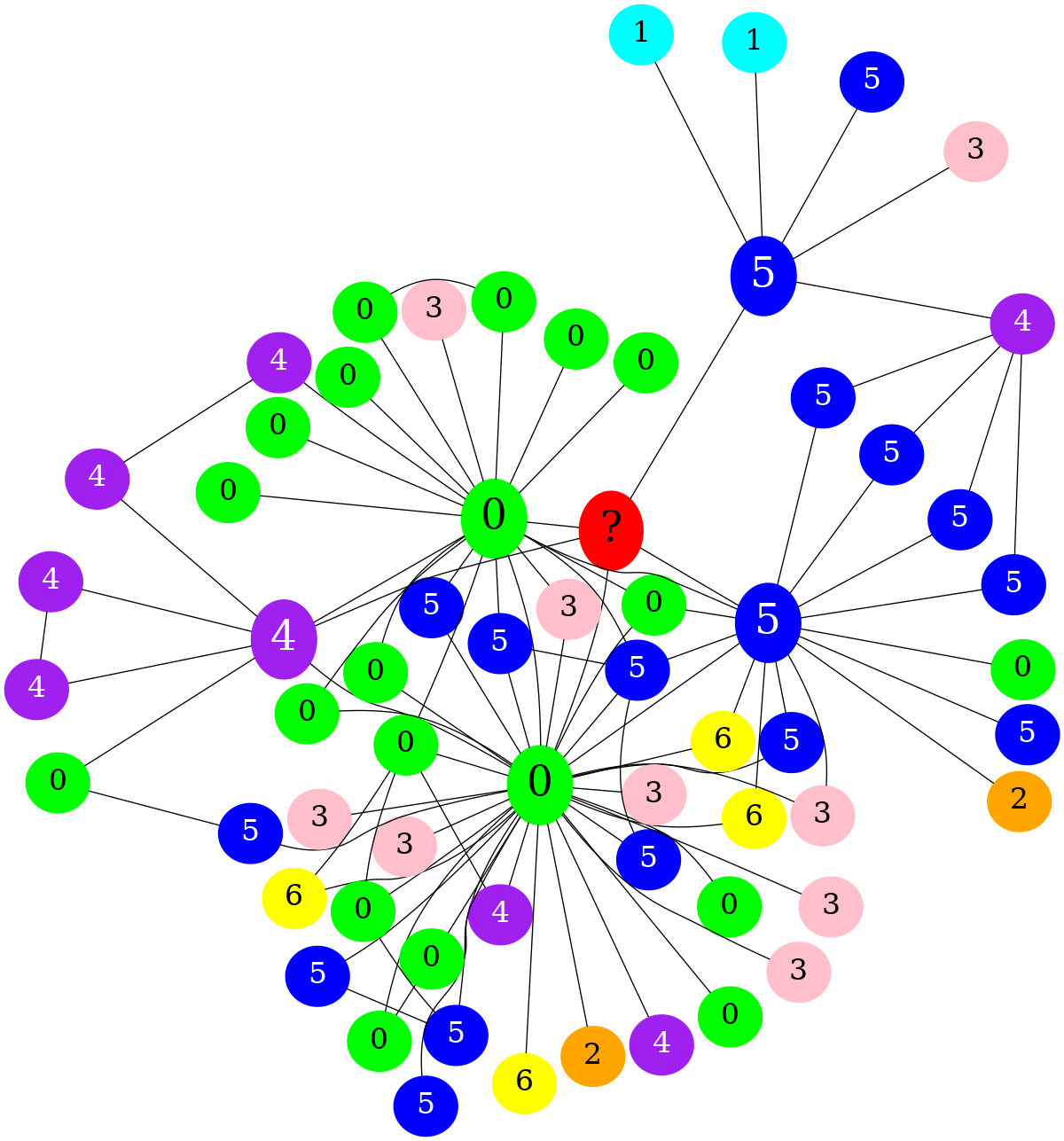}
    \caption{(Ground = 0) Without additional context or rules for how labels are assigned, it is not possible to accurately predict the label of the red node.}
     \end{subfigure}
     \hfill
     \begin{subfigure}[b]{0.25\textwidth}
         \centering
         \includegraphics[width=\columnwidth]{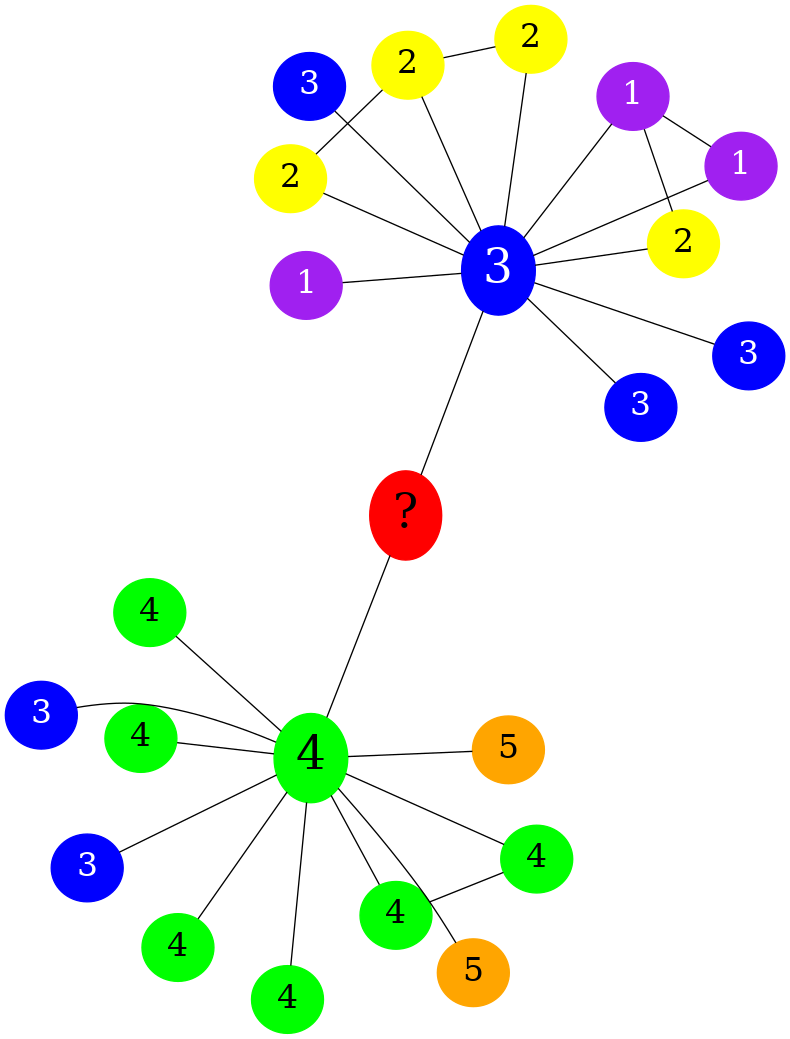}
         \caption{(Ground = 4) The label cannot be determined with certainty due to the lack of a discernible pattern or rule that associates a node's color or its connections with its label.}
     \end{subfigure}
     \hfill
     \begin{subfigure}[b]{0.25\textwidth}
         \centering
         \includegraphics[width=\columnwidth]{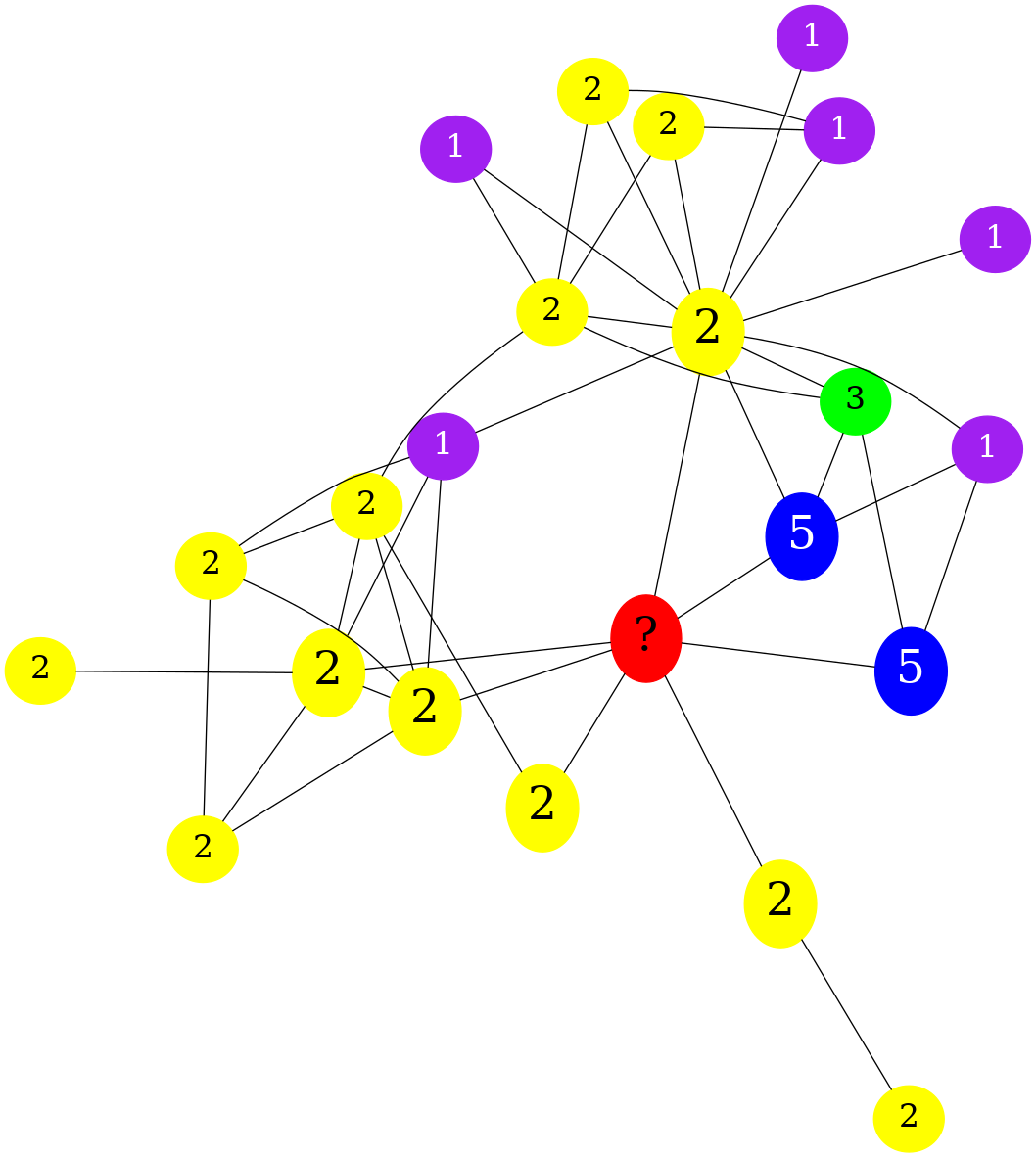}
         \caption{(Ground = 2) The label cannot be determined with certainty due to the lack of a clear pattern in the graph and no previous examples of red nodes to infer from. }
     \end{subfigure}
\caption{Examples of graphs where VLM (GPT-4V) returned -1 or denied to predict a label and the reason for denial. The ground truth for this graph is given in brackets. This \textit{highlights the need to clarify labeling strategies and few shot learning.}}
\label{tab:denial_cases}
\end{figure*}

\noindent \textbf{Qualitative analysis of denial of classification in the Image Modality}
Figure \ref{tab:denial_cases} shows instances from multiple datasets where GPT-4V, using image modality, did not assign labels (returned $-1$) to graph nodes, explaining the reasons for denial. Key observations include: a) the LLM lacked \textit{explicit context on label assignments} to nodes, as the encoding only implicitly indicated labels through node colors, with red reserved for unlabeled nodes. b) For one image, the absence of a clear \textit{link between node colors and labels}, exacerbated by high heterophily, caused confusion. c) Another case highlighted the need for \textit{few-shot learning}, suggesting that showing the LLM similar graph examples could help it learn to identify unlabeled (red) nodes more accurately. 

\noindent \textbf{Insights from GraphTMI}
In our evaluation of node classification accuracy using the \textsc{GraphTMI} benchmark across various modalities, we found in Figure \ref{fig:fig9} that ``easy'' tasks (characterized by high homophily and simpler structures) showed comparable accuracy across text, image, and motif modalities. However, for ``medium'' or ``hard'' tasks, marked by heterophyllous nature or complex structures, the image modality outperformed others, followed by the motif modality, underscoring the importance of global information in LLM processing. Notably, ``hard'' graphs achieved the highest accuracy with the motif modality, indicating the value of balancing local and global information. This suggests a \textit{growing effectiveness of image and motif modalities in enhancing graph reasoning tasks like node classification}.

\subsection{Modality Specific Results}
\begin{figure*}[ht!]
    \centering
    \begin{minipage}{0.48\textwidth}
        \centering
        \includegraphics[width=0.9\linewidth]{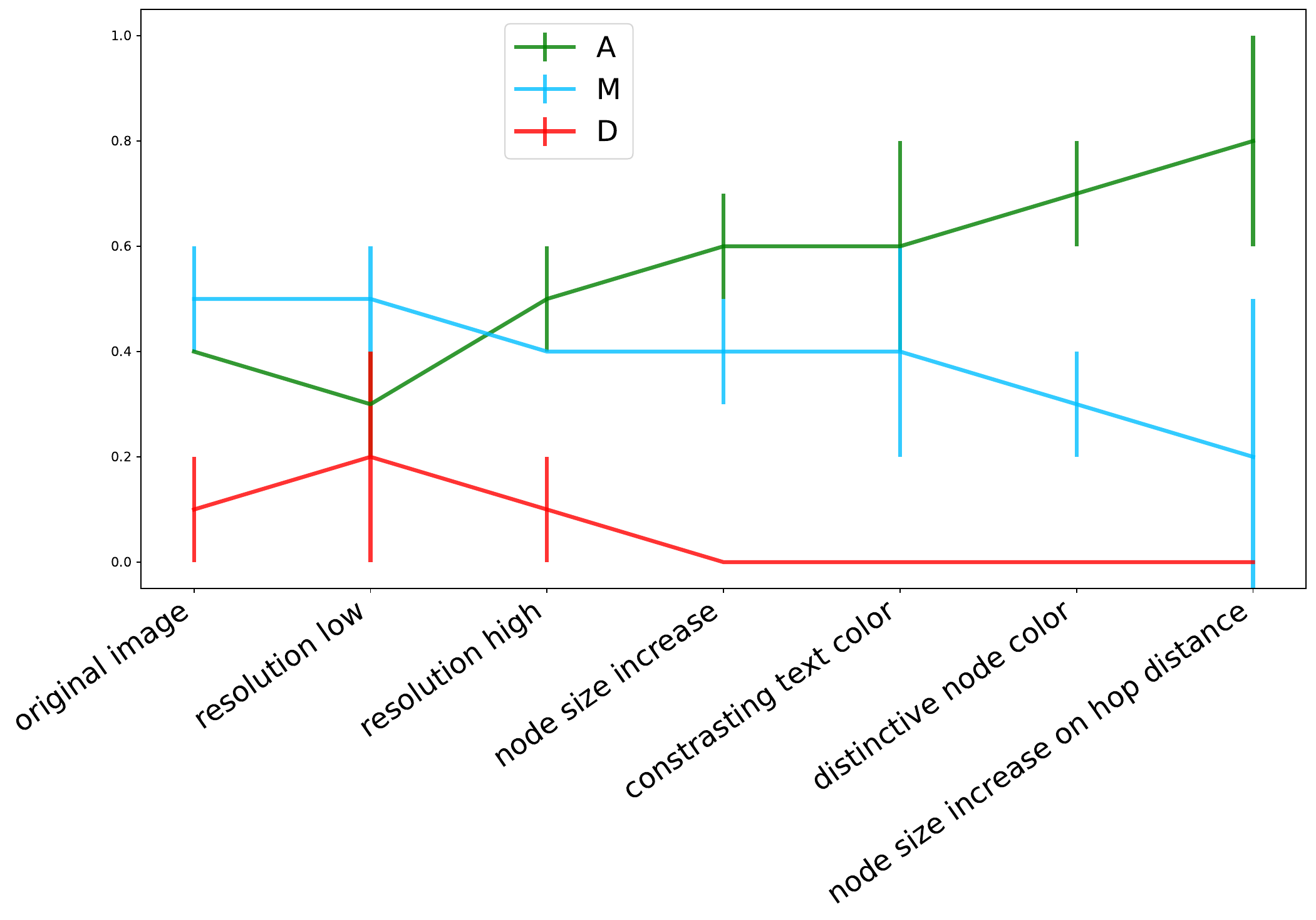}
        \caption{Our comparison of image representations (x-axis) with mean metrics (y-axis) shows that \textit{human readability of images correlates with classification performance}, considering Accuracy Rate (A $\uparrow$), Mismatch Rate (M $\downarrow$), and Denial Rate (D $\downarrow$), with desired trends indicated in brackets.}
        \label{fig:fig7}
    \end{minipage}\hfill
    \begin{minipage}{0.48\textwidth}
        \centering
        \includegraphics[width=0.95\linewidth]{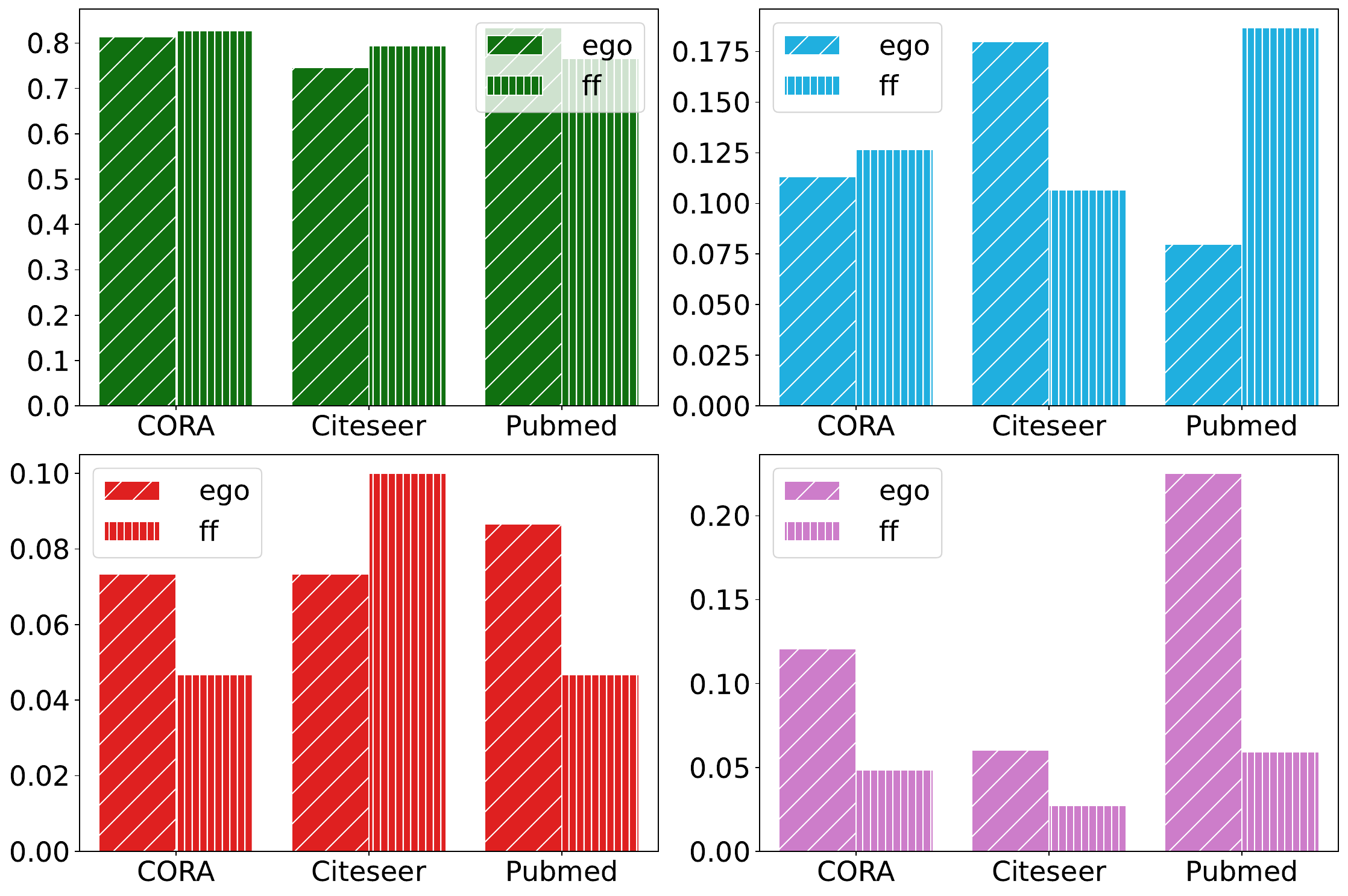}
        \caption{Each dataset and sampling type (x-axis) is mapped against mean metrics (y-axis), with bar textures distinguishing between ego graph (ego) and forest fire (ff) sampling; the metrics include \textcolor{green}{accuracy rate} ($\uparrow$), \textcolor{blue}{ mismatch rate} ($\downarrow$), \textcolor{red}{denial rate} ($\downarrow$), and \textcolor{purple}{token limit fraction} ($\downarrow$), indicating the desired trends for each. \textit{Graph structures of different datasets and sampling strategies influence node classification performance.}}
        \label{fig:fig5}
    \end{minipage}
\end{figure*}

\noindent \textbf{Text modality results}: 
Figure \ref{fig:fig4} shows that using the \textit{Adjacency List} as the mode of edge representation with node label mapping is the \textit{most informative encoding function}, which balances the trade-off between high accuracy and low token limit fraction. Figure \ref{fig:fig5} shows how metrics vary across datasets with different graph structures and sampling strategies. For \textsc{CORA}, a small, dense, and clustered graph, both sampling methods yield high accuracy, with forest fire (ff) sampling resulting in a lower denial rate. \textsc{Citeseer}, with its local clustering nature, struggles with ff sampling, showing the highest denial rate and the lowest mean accuracy, indicating difficulty in accurate predictions. In contrast, large and highly connected \textsc{Pubmed} generates larger samples through ego graph sampling, leading to higher token limit fractions. \textsc{Citeseer}'s fragmented, disconnected nature results in smaller ego graph samples and lower token limit fractions. Thus, we can see \textit{graph structure and sampling strategy significantly impact performance metrics}.

\noindent \textbf{Motif modality results}: Figure \ref{fig:fig6} shows GPT-4's \textit{improved performance by adding the ``triangle and star attached to $?$ node'' motif} in the motif modality encoder (detailed in Appendix Table \ref{tab:motif_representations}). This enhancement in mean accuracy and other metrics is attributed to the effective combination of local and global context provided to the LLM through node-label mapping and the associations within triads or star motifs.

\noindent \textbf{Image modality results}: Figure \ref{tab:img_modality_changes} shows different tweaks to image representation, and Figure \ref{fig:fig7} demonstrates that optimal node classification correlates with high accuracy and low denial and mismatch rates. Interestingly, \textit{as human image readability increases, metric performance also improves, highlighting the easier use of images over text} for LLM prompts.


%% file: tex/related_work.tex
\noindent \textbf{LLMs with Graphs}: Graph Neural Networks (GNNs) are renowned for their effectiveness in node classification and link prediction \cite{dwivedi2020benchmarking}, with applications in diverse fields like social networks, computer vision, and biology \cite{hou2022measuring}. GNNs struggle with processing non-numeric data like text and images, necessitating preprocessing such as feature engineering \cite{wang2021bag}. In contrast, recent studies have explored using Large Language Models (LLMs) for graph reasoning, demonstrating their potential in complex tasks \cite{huang2022language}. This includes using LLMs for feature enhancement \cite{chen2023exploring}, node classification \cite{chen2023exploring}, and training neural networks in graph-based tasks \cite{he2023explanations}, with benchmarks like NLGraph \cite{wang2023can} assessing LLMs in traditional graph challenges. These studies typically employ LLMs as sub-components within graph learning frameworks. \textit {Our research examines LLMs' ability to process graph modalities directly, aiming to understand LLMs' intrinsic graph-handling capabilities, thus presenting a novel direction in the field.} 
\noindent \textbf{Prompt Design for Graphs}: Prompting strategies for querying large language models (LLMs) aim to optimize the prompt text for enhanced task performance.  Few-shot in-context learning \cite{brown2020language} provides examples with desired outputs for the model to learn and generalize. Chain-of-thought (CoT) prompting \cite{wei2022chain} offers step-by-step problem-solving examples, leading the model to develop reasoning paths, while its zero-shot variant \cite{kojima2022large} initiates reasoning with a starter phrase. Bag prompting \cite{wang2023can} focuses on graph tasks, recommending graph construction before the task. Format explanations and role prompting \cite{guo2023gpt4graph} are proposed for better task clarity and strategic input organization to leverage LLMs' learning capabilities. Self-prompting involves the LLM refining prompts via context summarization, tackling issues with complex or insufficient graph data. \textit {Our study employs zero-shot prompting, providing only a task description to the LLM, to concentrate on the impact of modalities without the influence of varied prompt designs.}



%% file: tex/conclusion.tex
This study explores the application of LLMs in graph-structured data, evaluating their strengths and weaknesses in node classification using various input modalities like motif and image for effective data representation. Introducing the \textsc{GraphTMI} benchmark highlights the image modality's efficiency in token limit management. Although LLMs have progressed in graph data processing, they still don't match the performance of GNNs in practical settings. The research advocates for future work combining different modalities to improve node classification, combining LLM-based methods with GNNs, applying these techniques to complex, text-dense graphs, and delving into link prediction and community detection to broaden applications and insights across multiple domains.

%% file: tex/futurework.tex




%% file: tex/limitations.tex
Although innovative in applying LLMs to graph-structured data, this research faces key limitations. The computational demands of detecting network motifs, essential for understanding complex network dynamics, pose a significant challenge. This process requires extensive computational power and advanced algorithms, limiting scalability and efficiency. We subvert these challenges by restricting our subgraph sample size to 3 hops of an ego graph. Additionally, the study is constrained by the current rate limitations of GPT-V, which affects its ability to process and analyze data at scale efficiently. This limitation restricts the depth and scope of the analysis, although future advancements in GPT-V may mitigate this issue. A further limitation lies in the study’s simplistic approach to estimating homophily, relying merely on label count and neglecting the importance of hop distance. This overlooks critical network structure and node similarity aspects, leading to a potentially oversimplified analysis. Incorporating hop distance could provide a more accurate representation of network homophily. These limitations underscore the need for further advancements in computational techniques, model capabilities, and more nuanced theoretical methods in network analysis.

%% file: tex/appendix.tex
\begin{table*}[ht!]
\scriptsize
\begin{tabular}{m{0.20\linewidth}p{0.10\linewidth}p{0.29\linewidth}p{0.26\linewidth}p{0.10\linewidth}}
\hline
\textbf{Image} & \textbf{Ground Truth} & \textbf{Text\_Response} & \textbf{Text + Image\_Reponse} & \textbf{Image\_Response}\\
\hline
\raisebox{-\totalheight}{\includegraphics[width=\linewidth]{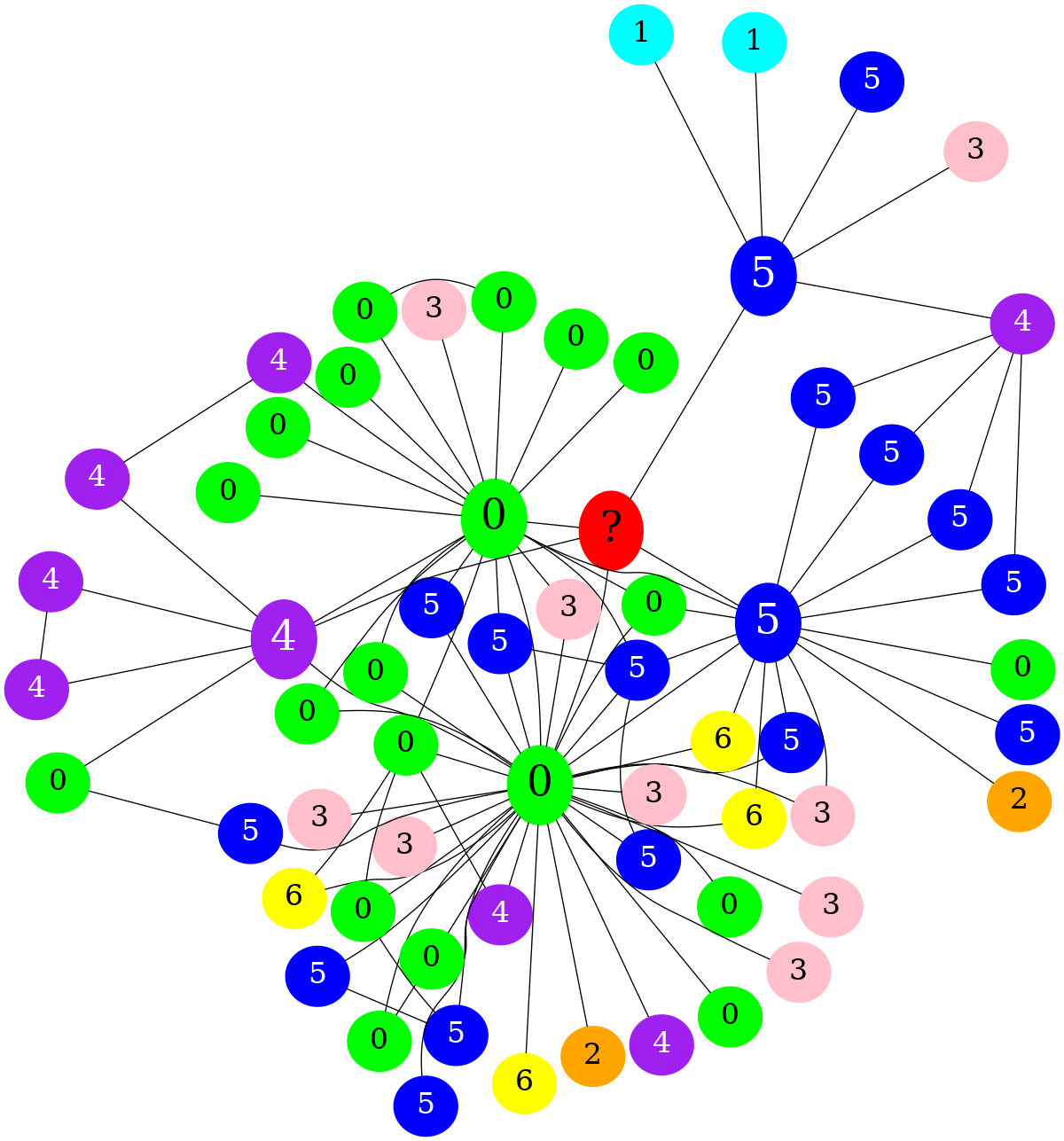}} & \textit{Label of Node = 5}  & The label of a node can often be predicted based on the labels of its neighboring nodes. In this case, the node in question is '1677', and its neighbors are '88', '415', '1309', and '191', '1908'. There is no clear majority label among these neighbors. Therefore, it is not possible to predict the label of node '1677' based on this information. \textit{Label of Node = -1}  & From the adjacency list, we can determine that the neighbors of node "1677". Given that node "1677" is connected to two nodes with label 0, two nodes with label 5, and one node with label 4, we can attempt to predict its label based on the labels of its neighbors. Since nodes "1309" and "191" both have the label 5 and are each directly connected to "1677," it's reasonable to refer to the homophily in networks concept where nodes are likely to be of the same type (label) as their neighbors. Based on this information, it seems plausible that the red node "1677" would also have the label that appears most frequently among its neighboring nodes, which is label 5. \textit{Label of Node = 5}  & \textit{Label of Node = 0} \\
\hline
\raisebox{-\totalheight}{\includegraphics[width=\linewidth]{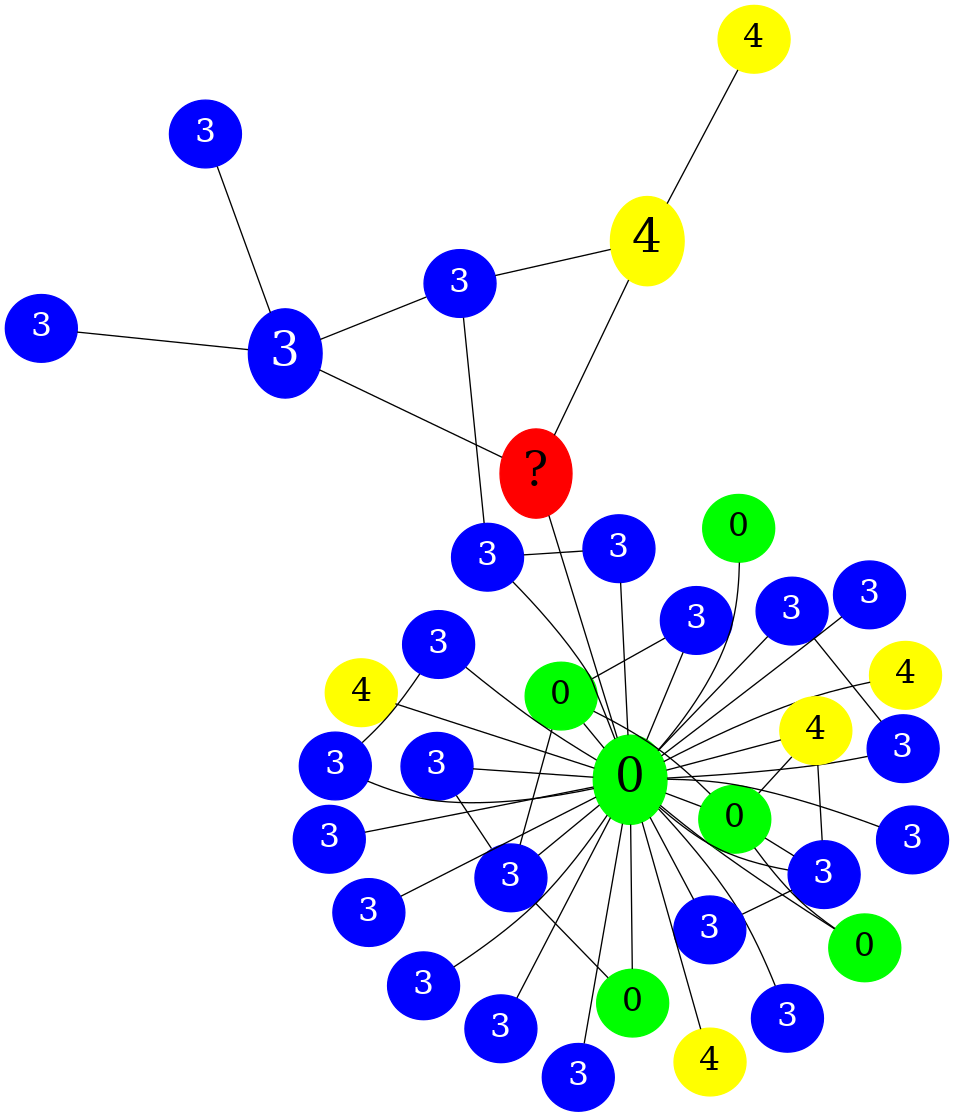}} & \textit{Label of Node = 3}  & The node 1329 is connected to nodes 733, 994, and 78. The labels of these nodes are 0, 3, and 4 respectively. Since there is no majority label among these, the label of node 1329 cannot be determined. \textit{Label of Node = -1} & \textit{Label of Node = 3} & \textit{Label of Node = 3} \\
\hline
\raisebox{-\totalheight}{\includegraphics[width=\linewidth]{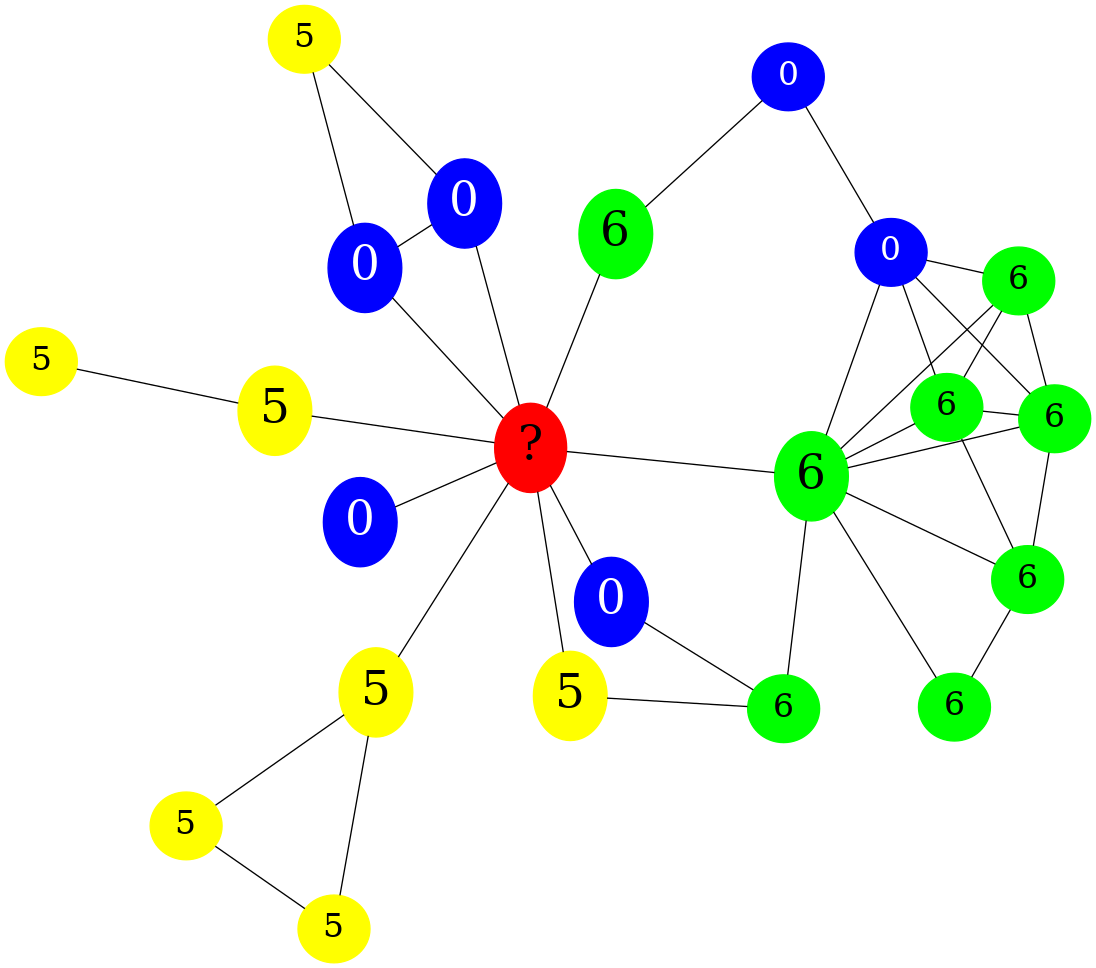}} & \textit{Label of Node = 6}  & The node 702 is connected to nodes 777, 779, 2069, 2070, 2101, 822, 993, 1368, and 1637. The labels of these nodes are 0, 6, 0, 0, 5, 5, 6, 5, and 0 respectively. The most common label among these is 0. Therefore, we can predict that the label of node 702 is also 0.\textit{Label of Node = 0} & \textit{Label of Node = 6}  & \textit{Label of Node = 6} \\
\hline
\raisebox{-\totalheight}{\includegraphics[width=\linewidth]{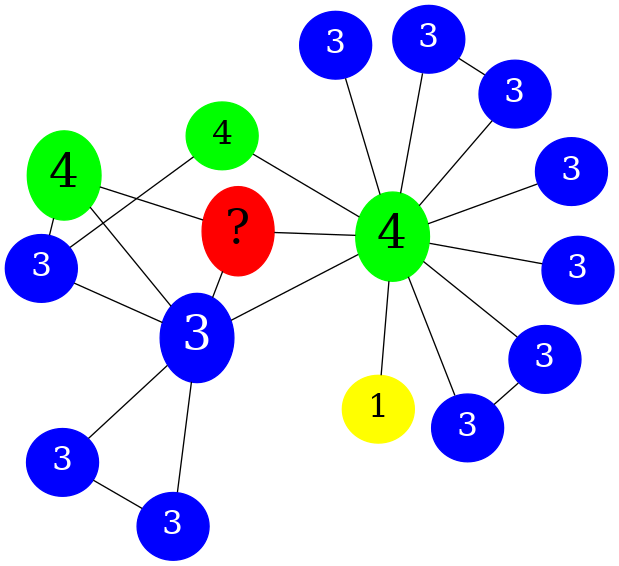}} & \textit{Label of Node = 3}  & The node 141 is connected to nodes 1002, 2034, and 740. The labels of these nodes are 3, 4, and 4 respectively. Since the majority of the connected nodes have the label 4, we can predict that the label of node 141 is also 4. \textit{Label of Node = 4} & \textit{Label of Node = 3}  & \textit{Label of Node = 3} \\
\hline
\end{tabular}
\caption{The table shows the GPT-4 and GPT4-V responses with the text modality, text+image modality, and image modality. \textit{We observe that on combining modalities, the label originally classified as -1 is correctly classified (first two rows), or the original misclassification is rectified (last two rows).} }
\label{tab:qual}
\end{table*}


\section {LLM Experiments}
\label{sec:A}
\subsection{Qualitative Analysis on Combining Modalities}
We perform a qualitative analysis of the response returned by LLMs by utilizing the text, image, and text combined with image encoding modalities. The intuition here is that the local context provided by the text modality might not be enough for some predictions and could be supplemented through the global context provided by the image modality. \textit{Table \ref{tab:qual} illustrates that misclassifications and denials by the LLM using text modality could be rectified by using the image modality}. For the first two rows, $-1$ classifications or LLM denials are changed to the correct classification on incorporating the global context of the image modality. We can see in the response that the notion of ``homophily'' is clearer to the VLM in the image modality. For the last two rows, we see that the graph is originally misclassified, but then this is corrected by incorporating the image modality. We make similar observations on combining text and motif modalities, and this could be because another factor important to node classification is the presence of motifs, which is highlighted through the motif modality. 
\subsection{Comparing encoding modalities for different datasets and sampling}
\begin{figure*}[ht!]
    \centering
    \begin{minipage}{0.40\textwidth}
        \centering
        \includegraphics[width=\linewidth]{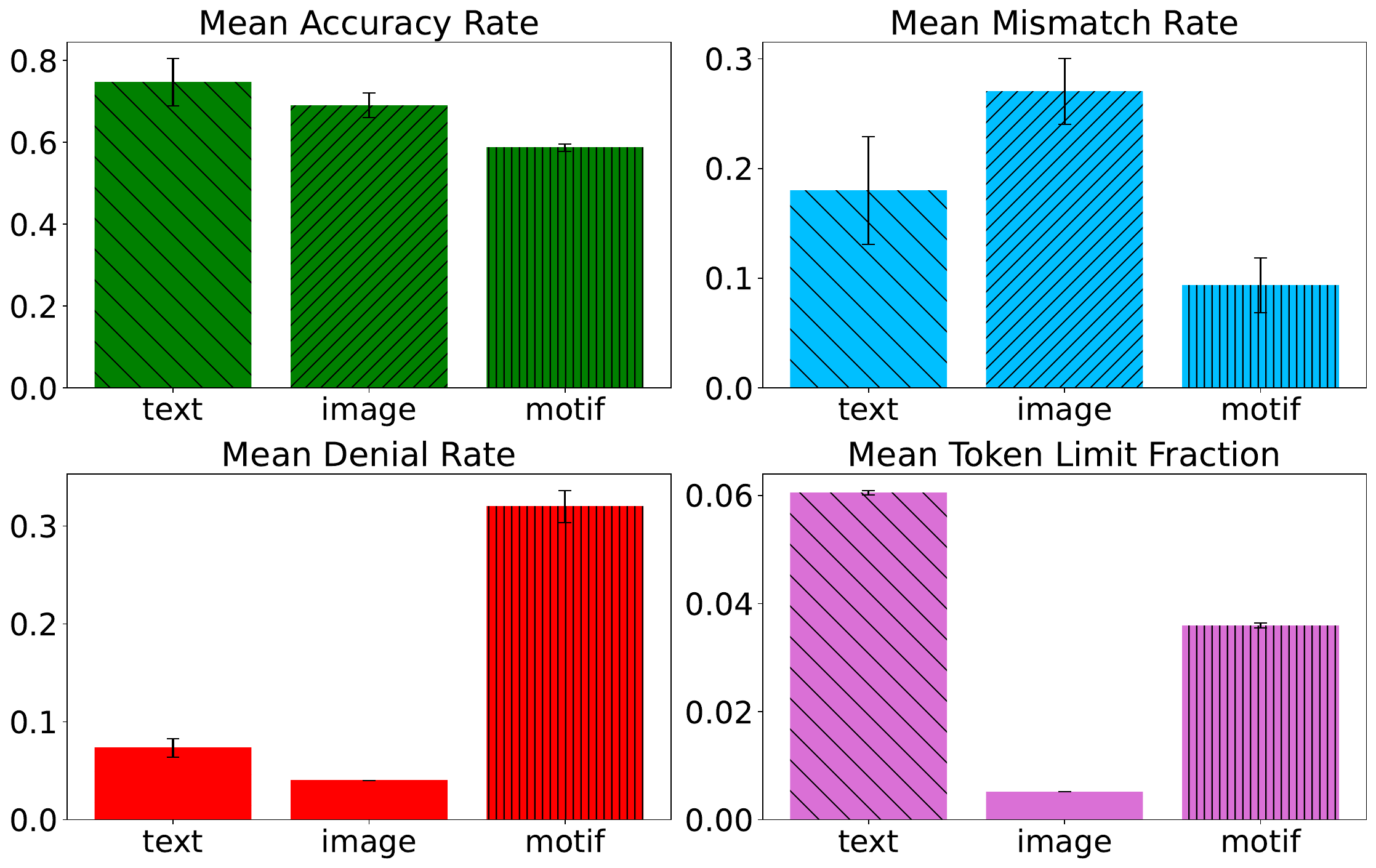}
        \caption{Citeseer with ego graph sampling}
        \label{fig:image1}
    \end{minipage}
    \hfill 
    \begin{minipage}{0.40\textwidth}
        \centering
        \includegraphics[width=\linewidth]{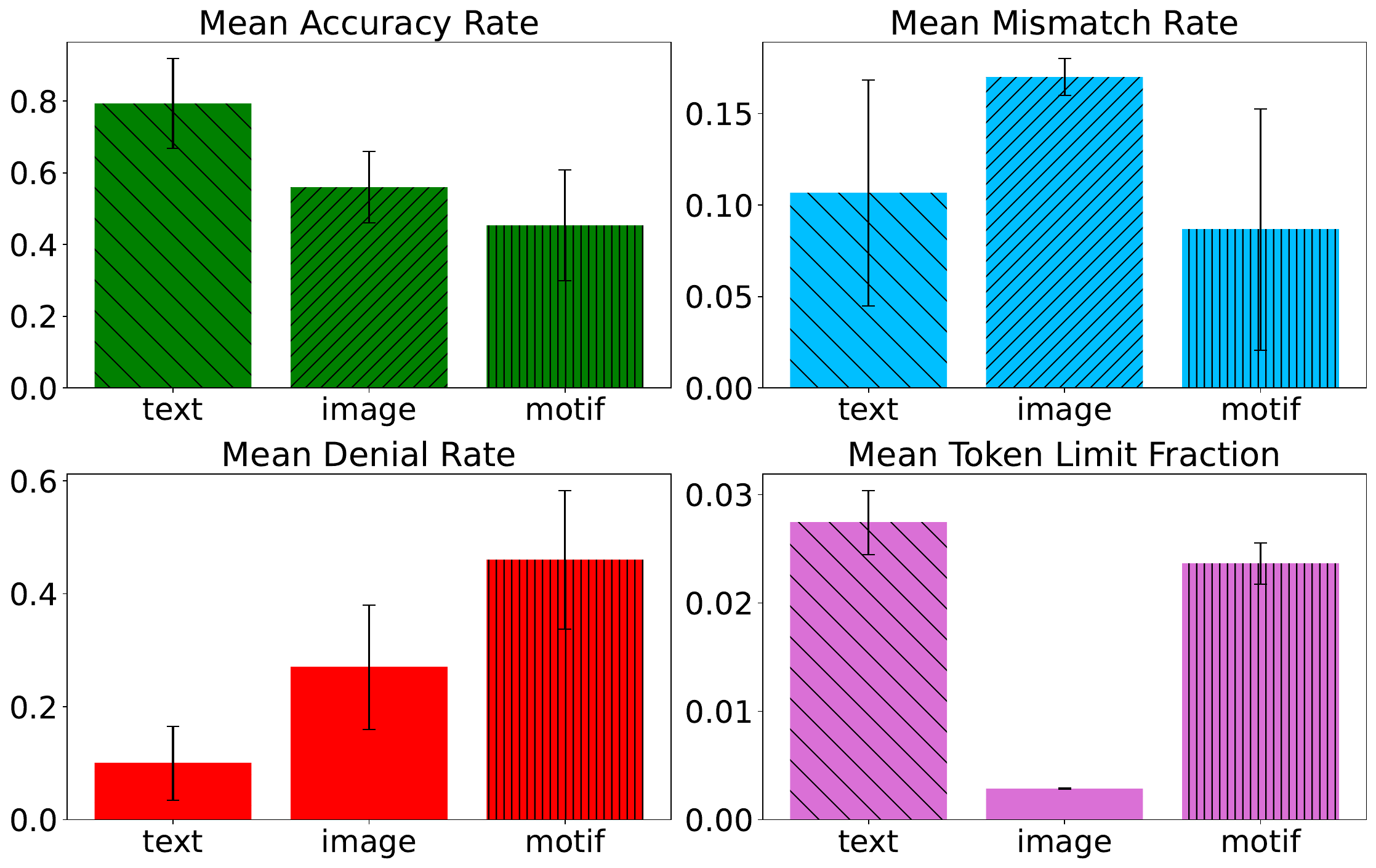}
        \caption{Citeseer with forest fire sampling}
        \label{fig:image2}
    \end{minipage}
    \hfill
    \begin{minipage}{0.40\textwidth}
        \centering
        \includegraphics[width=\linewidth]{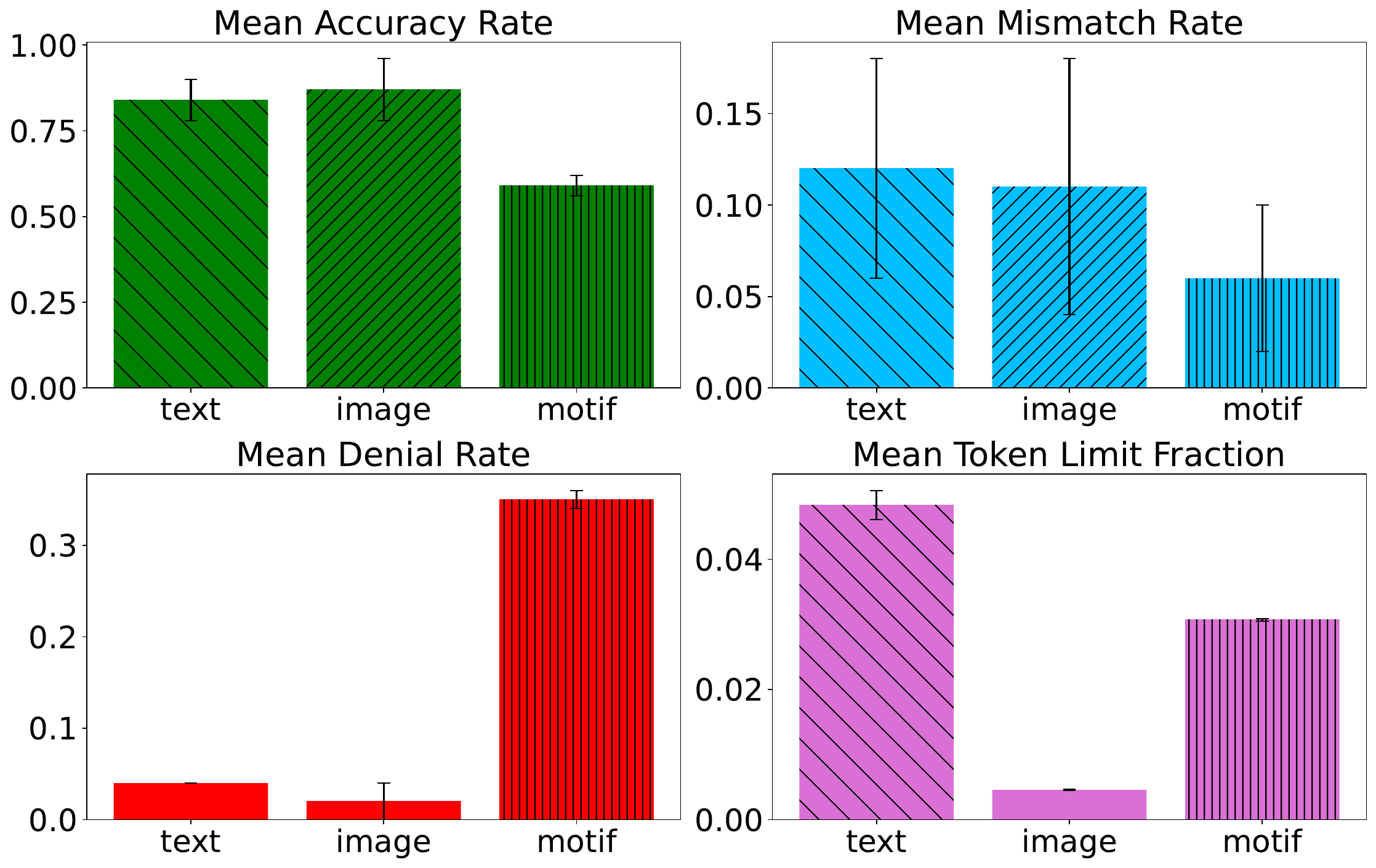}
        \caption{Cora with forest fire sampling}
        \label{fig:image3}
    \end{minipage}

    \begin{minipage}{0.40\textwidth}
        \centering
        \includegraphics[width=\linewidth]{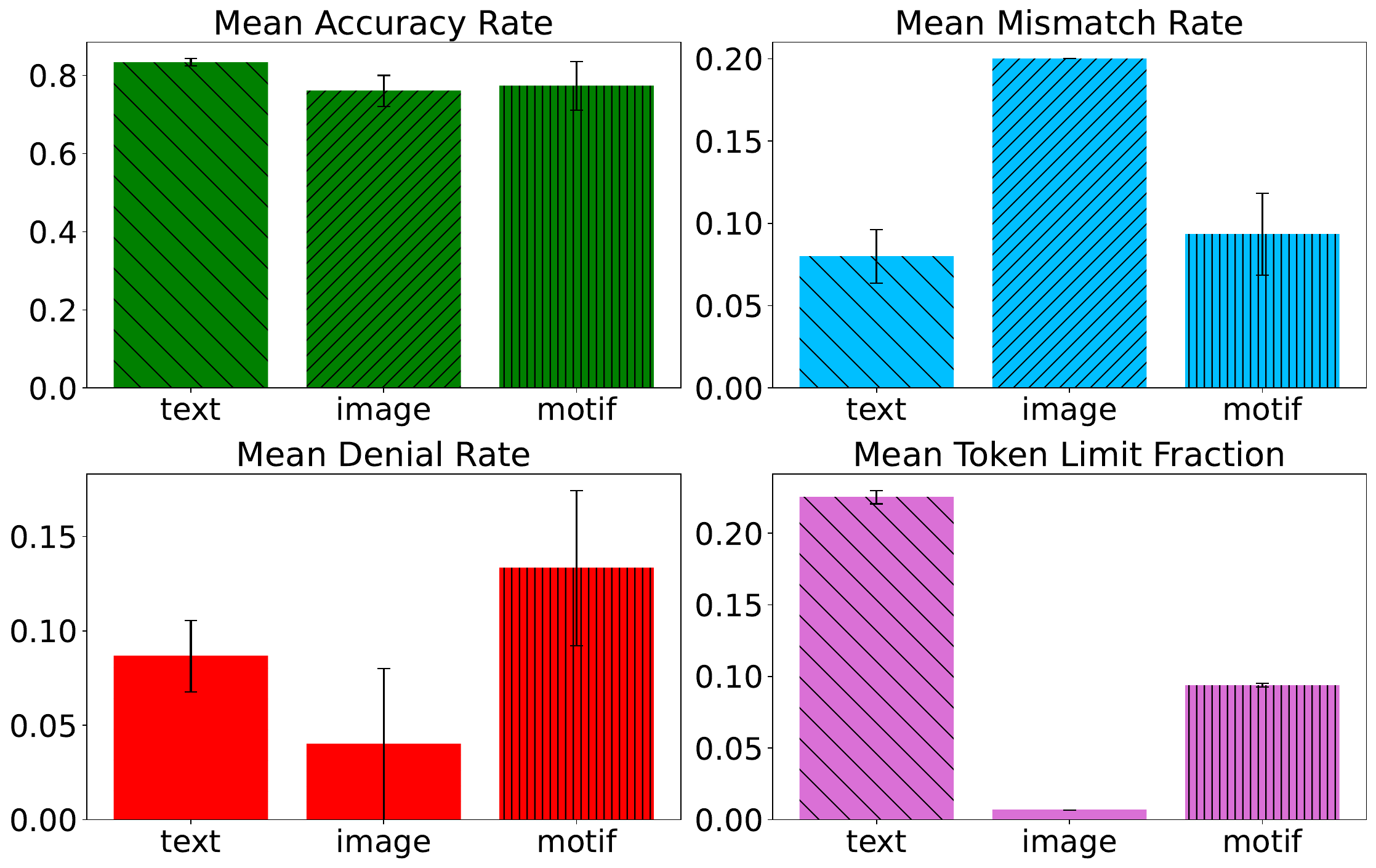}
        \caption{Pubmed with ego graph sampling}
        \label{fig:image4}
    \end{minipage}
    \hfill
    \begin{minipage}{0.40\textwidth}
        \centering
        \includegraphics[width=\linewidth]{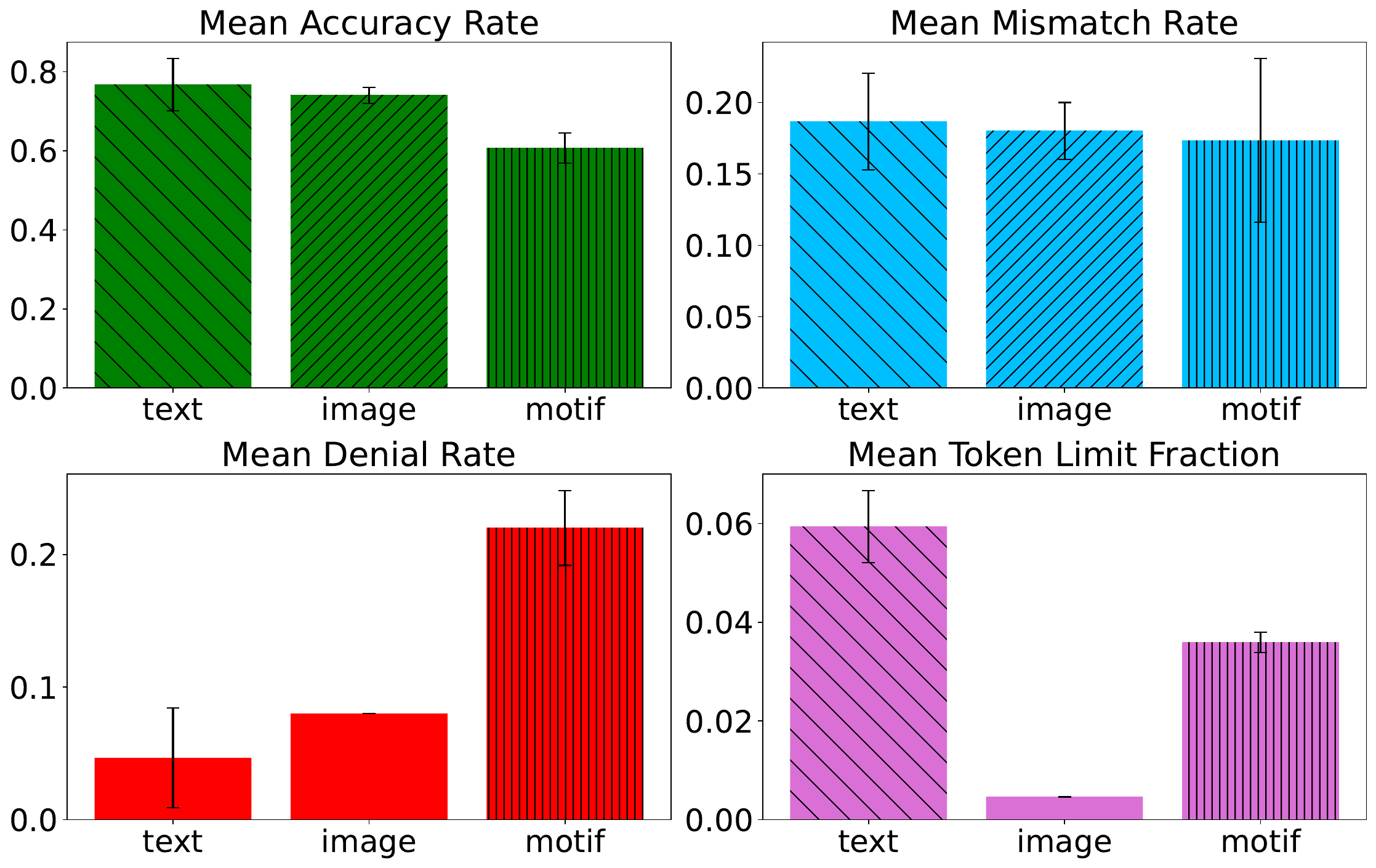}
        \caption{Pubmed with forest fire sampling}
        \label{fig:image5}
    \end{minipage}
    \caption{Modality comparison (text, motif, and image) with the graph structure and sampling type shows the \textit{clear dependency of graph structure and sampling on node classification performance}.}
    \label {fig:modality_datasets}
\end{figure*}

Our study evaluates various encoding modalities — text, motif, and image — with ego graphs from CORA detailed in the main manuscript. Figure \ref{fig:modality_datasets} extends this analysis to the other datasets and sampling techniques. The findings corroborate our assertion that graph structure and the chosen sampling method significantly influence node classification outcomes. Particularly, samples derived from the forest fire method, which emphasize the global configuration while being sparser and less connected than ego graphs, exhibit increased misclassification rates when using the image modality due to limited information for accurate inference and greater instances of non-committal predictions with the motif modality due to the absence of a discernible overarching structure.  

\subsection{Graph TMI Benchmark}
We decide on graph ``difficulty'' based on the dual criteria of 1) count of motifs and 2) homophily in the graph. We apply a naive heuristic to decide homophily, i.e., the count of the distinct labels in the graph. If the count of distinct labels $<3$, the graph is considered \textit{easy}. If the count is $\geq 3$ and $<5$, it is considered \textit{medium}, and if the count is $\geq 5$, it is considered \textit{hard}. To decide the motif criteria, we count the total number of motifs( focusing on just triads, star motifs, and cliques) in a graph. For example, if this count of motifs $\leq 10$ for the CORA dataset, the graph is considered \textit{easy}. If the count is $>10$ and $\leq 20$, it is regarded as \textit{medium}; if the count is $>20$, it is considered \textit{hard}. Some graphs can have both the homophily and motif criteria applicable to them; for instance, Figure \ref{tab:task_hierarchy} (b) can be classified as \textit{medium} based on homophily, \textit{hard} based on the count of motifs. This leads us to combine the homophily and the count of motif criteria to define the ``task difficulty''. Thus, we can have $2^3 = 8$ categories of difficulty, and the final difficulty label is decided by choosing the higher annotation between homophily and count of motif classification). So, a graph with criteria \{easy, hard\} will be assigned the final task difficulty, \textit{hard}. Thus, we can classify graphs based on our ``task difficulty'' heuristic, and we introduce \textsc{GraphTMI} (Graph Text-Motif-Image), a novel benchmark dataset of input graph structures paired with their associated modality encodings.
\begin{table}[H]
\centering
\footnotesize 
\setlength{\tabcolsep}{4pt} 
\begin{tabular}{lccc}
\toprule
  & \textbf{homophily} & \textbf{motif} & \textbf{count} \\
\midrule
0 & easy & easy & 7 \\
1 & easy & hard & 4 \\
2 & easy & medium & 6 \\
3 & hard & easy & 1 \\
4 & hard & hard & 8 \\
5 & hard & medium & 2 \\
6 & medium & easy & 5 \\
7 & medium & hard & 5 \\
8 & medium & medium & 12 \\
\bottomrule
\end{tabular}
\caption{Statistics about the number of graphs classified as easy, medium, or hard through the homophily and the number of motifs criteria. All possible combinations are covered in our benchmark ($3^2 = 9$). } 
\label{tab:homo_motif}
\end{table}
\begin{table}[H]
\centering
\footnotesize 
\setlength{\tabcolsep}{6pt} 
\begin{tabular}{lcc}
\toprule
  & \textbf{difficulty} & \textbf{count} \\
\midrule
0 & easy & 7 \\
1 & hard & 20 \\
2 & medium & 23 \\
\bottomrule
\end{tabular}
\caption{Statistics about the number of problems finally classified as easy, medium, or hard based on task difficulty, a function of homophily, and number of motifs.} 
\label{tab:task_diff}
\end{table}

\subsection{Modality Specific Experiment Details}
\noindent \textbf{Token limits for each modality}
Due to their architecture, transformer-based models like GPT-3 and GPT-4 have a fixed-size attention window. This determines how many tokens the model can ``remember'' or pay attention to at once. This limit also manages the computational cost of running the model and the model's performance. The token limit constraint for GPT-4 is 8192 tokens, while for GPT-4V(vision), the limit is claimed to be 128000 tokens, but currently, only the preview version has been released, and the actual limit is 10000 tokens.

\noindent \textbf{Rate limits for each LLM}
The rate limit for GPT-4 is 10K RPM (requests per minute), and for GPT-4V, the rate limit is 100 RPD (requests per day). 

\noindent \textbf{Modality Experiment Parameters}
For all modality types, we sample 50 graphs for all datasets, the number of hops considered = 3, no of runs = 2, and perform ego graph and forest fire sampling. We report the mean and standard deviation directly or through error bars in the visualization for all metrics. In the paper, we report the results from ego graph sampling because node classification typically needs a localized view around specific nodes, best provided by ego graph sampling. 

\subsubsection {Text Modality}
Encoding graphs as text can be separated into two key parts: First, the mapping of nodes to their corresponding labels in the graph, and second, the encoding of edges between the nodes. We encode the node-to-label mapping as a dictionary of type \{node ID: node label\}. Finding a concise yet informative representation of the graph structure and edge representation is essential. 
Example of a prompt using text modality : 

\begin{tcolorbox}[colback=mycolor,boxsep=1pt,left=4pt,right=4pt,top=4pt,bottom=4pt]
\small
\noindent\textbf{Task}: Node Label Prediction (Predict the label of the node marked with a ?) given the adjacency list information as a dictionary of type ``node: neighborhood'' and node-label mapping in the text enclosed in triple backticks. The response should be in the format ``Label of Node = \textless predicted label\textgreater''. If the predicted label cannot be determined, return ``Label of Node = -1''.\\
\textasciigrave \textasciigrave \textasciigrave 
\textbf{AdjList}: \{1: [2,3], 2: [3,4], 3: [1,2]\} \\
\textbf{Node-Label Mapping}: \{1: A, 2: B, 3: ?\} \textasciigrave \textasciigrave \textasciigrave
\end{tcolorbox}

\begin{table*}[ht]
\centering
\scriptsize
\begin{tabular}{@{}p{0.15\textwidth}p{0.45\textwidth}p{0.4\textwidth}@{}}
\toprule
\textbf{Edge Representation} & \textbf{Text Encoding} & \textbf{Description of Edge Representation} \\
\midrule
\textbf{Edgelist} &  Node to Label Mapping : 
Node 69025: Label 34| Node 17585: Label 10|...
\textit{Edge list: [(69025, 96211), (69025, 17585), (17585, 104598), (17585, 18844), (17585, 96211), (96211, 34515)]} &
An Edgelist is a graph data structure that represents a graph by listing the edge connections between two nodes. (A, B) indicates an connection between nodes A and B. \\
\midrule

\textbf{Edgetext} & Node to Label Mapping : 
Node 85328: Label 16| Node 158122: Label ?|...
\textit{Edge connections (source node - target node): Node 85328 is connected to Node 158122. Node 158122 is connected to Node 167226. } &
An Edgetext explicitly lists the connections between two nodes; for example, Node A is connected to Node B or Node A - Node B \\
\midrule
\textbf{Adjacency List} & Node to Label Mapping : Node 2339: Label 3| Node 2340: Label ?|...
\textit{Adjacency list: {1558: [2339, 2340], 2339: [1558, 2340], 2340: [2339, 1558]}} 
&
An adjacency list represents a graph as an array of linked lists. The index of the array represents a vertex, and each element in its linked list represents the other vertices that form an edge with the vertex. For example, A: [B, C] shows that A is connected to B and C. This gives an idea of node-neighborhood\\
\midrule
\textbf{GML} & 
GraphML: graph [\newline
  \hspace*{0.2cm}node [\newline
    \hspace*{0.4cm}id 2339\newline
    \hspace*{0.4cm}label 3\newline
  \hspace*{0.2cm}]\newline
  \hspace*{0.2cm}node [\newline
   \hspace*{0.4cm} id 2340\newline
   \hspace*{0.4cm} label ?\newline
  \hspace*{0.2cm}]\newline
  \hspace*{0.2cm}node [\newline
   \hspace*{0.4cm} id 1558\newline
   \hspace*{0.4cm} label 3\newline
  \hspace*{0.2cm}]\newline
  \hspace*{0.2cm}edge [\newline
  \hspace*{0.4cm}  source 2339\newline
   \hspace*{0.4cm} target 1558\newline
  \hspace*{0.2cm}]\newline
  \hspace*{0.2cm}edge [\newline
   \hspace*{0.4cm} source 2339\newline
   \hspace*{0.4cm} target 2340\newline
  \hspace*{0.2cm}]\newline
]\newline
 &
A GraphML format consists of an unordered sequence of node and edge elements enclosed within []. Each node element has a distinct id and label attribute contained within []. 
Each edge element has source and target attributes contained within [] that identify the endpoints of an edge by having the same value as the node id attributes of those endpoints. The node label information is embedded within the structure, meaning no node-label mapping is notneeded. \\
\midrule
\textbf{GraphML} & 
GraphML: <graphml xmlns=http://graphml.graphdrawing.org/xmlns xmlns:xsi=http://www.w3.org/2001/XMLSchema-instance \newline xsi:schemaLocation=http://graphml.graphdrawing.org/xmlns \newline  http://graphml.graphdrawing.org/xmlns/1.0/graphml.xsd>
  <graph edgedefault=undirected>\newline
  \hspace*{0.2cm}  <node id=2339 label=3 /> \newline
  \hspace*{0.2cm}  <node id=2340 label=? />\newline
  \hspace*{0.2cm}  <node id=1558 label=3 />\newline
  \hspace*{0.2cm}  <edge source=2339 target=1558 />\newline
  \hspace*{0.2cm}  <edge source=2339 target=2340 />\newline
  </graph>\newline
</graphml> &
A GraphML file consists of an XML file containing a graph element, within which is an unordered sequence of node and edge elements. Each node element should have a 
distinct id attribute as well as its label, and each edge element has source and target attributes that identify the endpoints of an edge by having the same value as the id attributes of those endpoints. The node label information is embedded within the structure meaning no node-label mapping is needed. \\

\bottomrule
\end{tabular}
\caption{Summary of edge representation passed as a part of the text modality encoder with their associated examples
and explanations. We find that the \textbf{Adjacency list} representation provides a granular yet not too verbose view of the graph being passed to the LLM.}
\label{tab:edge_representations}
\end{table*}

\begin{table}[ht]
\centering
\scriptsize 
\setlength{\tabcolsep}{3pt} 
\begin{tabular}{l|lll}
\toprule
& \textbf{CORA} & \textbf{Citeseer} & \textbf{Pub.med} \\
\midrule
Avg edges 2-hop & 62.70 ± 94.77 & 26.35 ± 61.70 & 129.36 ± 287.61 \\
\midrule
Avg nodes 2-hop & 36.78 ± 48.12 & 15.11 ± 24.73 & 60.05 ± 85.12 \\
\bottomrule
\end{tabular}
\caption{Subgraph Sampling Statistics about average number of nodes and edges in a 2-hop subgraph from each dataset.}
\label{tab:tab1}
\end{table}

\begin{figure}
  \centering
\includegraphics[width = 0.9\linewidth]{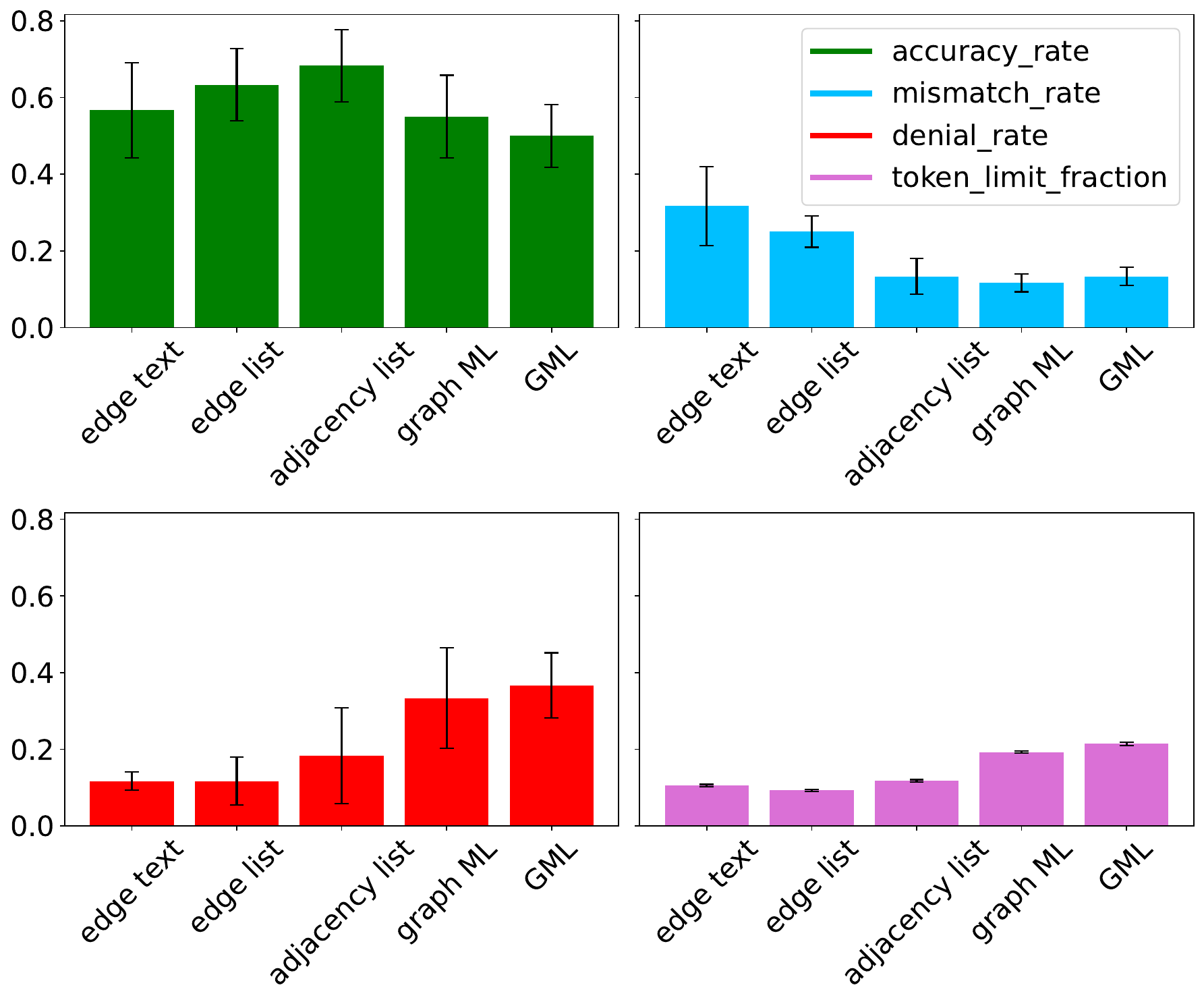}
  \caption{ We compare the edge representation type (x-axis) with the value of the mean metrics (y-axis). The desired trend is given in brackets for each metric. The highest performing edge representation is the ``adjacency list'' representation with the highest accuracy (A $\uparrow$) and low mismatch rate (M $\downarrow$)), denial rate (D $\downarrow$), and token limit fraction (T $\downarrow$).}
  \label{fig:fig4}
\end{figure}

\begin{table*}[ht]
\centering
\small
\caption{Graph Properties and Their Descriptions}
\label{tab:graph_properties}
\begin{tabular}{p{0.25\linewidth} p{0.7\linewidth}}
\toprule
\textbf{Name of Property} & \textbf{Description} \\
\midrule
Density & Measures how connected the graph is. It's the ratio of actual edges to possible edges. \\
\addlinespace
Degree Distribution & The distribution of node degrees. The histogram might follow a specific pattern (e.g., power-law distribution, Gaussian distribution). \\
\addlinespace
Average Degree & The average degree of nodes in the graph. \\
\addlinespace
Connected Components & A subgraph in which a path connects any two nodes. \\
\addlinespace
Clustering Coefficient & Measures the degree to which nodes tend to cluster together. \\
\addlinespace
Graph Diameter & The longest shortest path between any two nodes. It provides insight into the graph's overall size. \\
2hop nodes & Average number of nodes present in the subgraph at 2 hop distance from any node \\
\bottomrule
\end{tabular}
\end{table*}
\noindent \textbf{Impact of Edge encoding function}: Motivated by recent works \cite{fatemi2023talk,guo2023gpt4graph} describing the importance of selecting the appropriate text encoding for a graph, we experiment with different edge representations (Appendix Table \ref{tab:edge_representations}) on real-world datasets and evaluate the metrics for node classification and the results of this are illustrated in Figure \ref{fig:fig4}. \textit{``Adjacency list'' is the best-performing edge representation for the text modality.} 

\noindent \textbf{Impact of Graph Structure}: 
We selected diverse real-world citation datasets with unique network characteristics, as shown in Table \ref{tab:dataset_comparison}. These network properties are defined in Table \ref{tab:graph_properties}. The average number of nodes and edges in a 2-hop subgraph is also reported for CORA, Citeseer, and Pubmed datasets.

\noindent \textbf{Impact of Sampling Strategy}:
Graph sampling techniques are essential for applying LLMs in graph reasoning, particularly due to the limited context window of LLMs and the intricacy of real-world graphs \cite{wei2022evaluating}. Ego graph sampling centers on a specific node and its direct connections, forming a subgraph that mirrors these immediate relationships. In contrast, Forest Fire sampling randomly selects a node and expands from there, producing varying subgraphs in size and structure, influenced by factors like 'burning' probabilities. However, both methods have limitations and can potentially distort the overall structure of complex and extensive networks.

\subsubsection{Motif Modality}
\begin{figure}[ht!]
    \centering
        \centering
        \includegraphics[width=\linewidth]{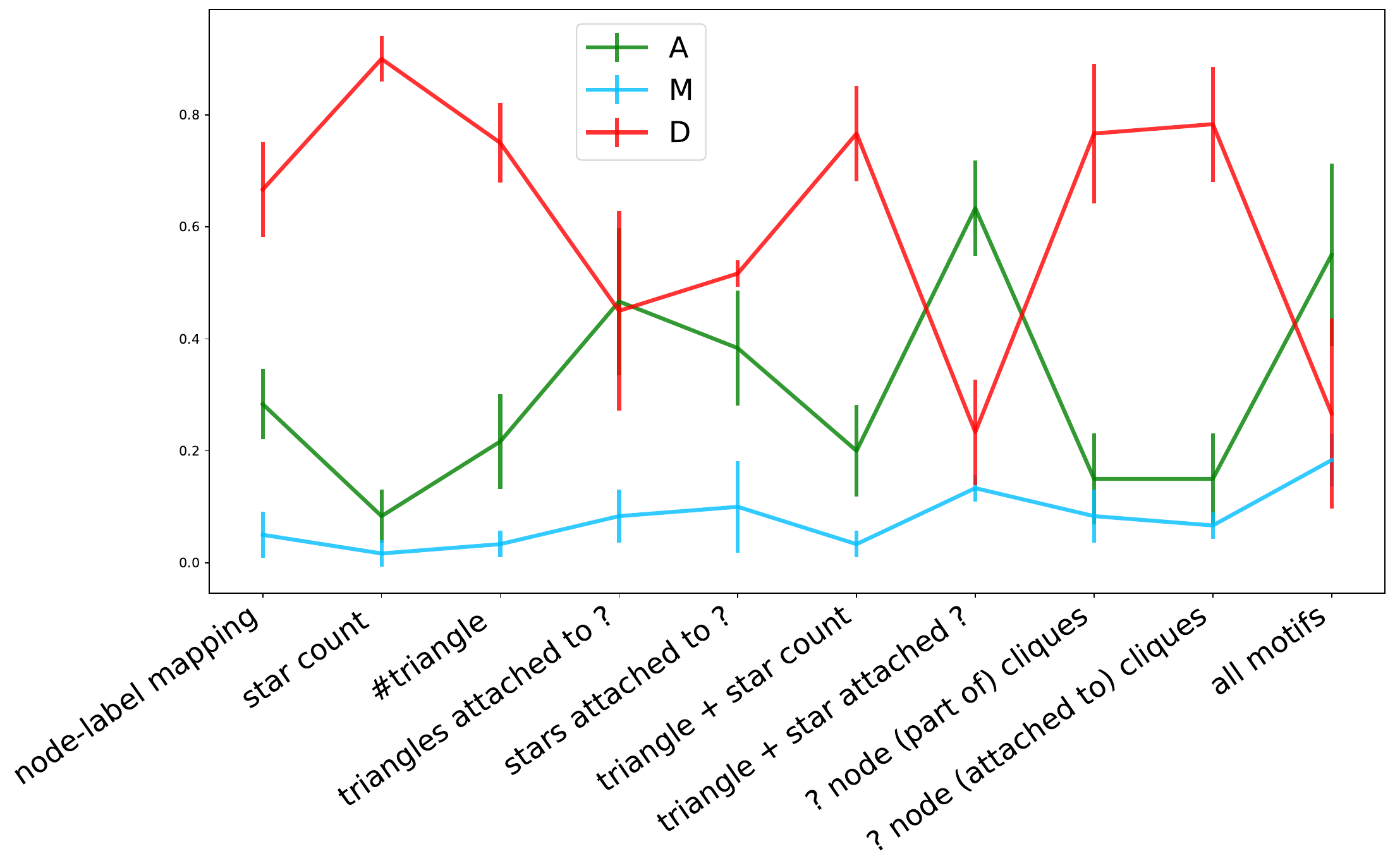}
        \caption{We compare the motif information (x-axis) to the mean metrics (y-axis). Desired trends are denoted in brackets. Metrics considered are Accuracy Rate (A $\uparrow$), Mismatch Rate (M $\downarrow$), and Denial Rate (D $\downarrow$). The highest performing motif information change ``triangle and star attached to $?$'' has higher accuracy and lower mismatch and denial rate.}
        \label{fig:fig6}
\end{figure}
Encoding graphs as motifs can be separated into two key parts: First, the encoding of nodes to their corresponding labels in the graph, and second, the motifs present around the $?$ (unlabeled) node. We encode the node-to-label mapping as a dictionary of type \{node ID: node label\}. We calculate motifs in the neighborhood of the $?$ nodes and pass this information to GPT-4 \cite{openai_gpt4} as the graph-motif information. Connections of an unlabeled node to significant nodes or groups (like stars or cliques) are more indicative of its label than just the count of graph motifs, with central nodes in star motifs or members of cliques heavily influenced by their neighbors' labels. We experiment with different network motifs as input to the modality encoder. Table \ref{tab:motif_representations} describes the different types of motifs considered, a description of the motif, and an example of the encoding generated as input to GPT-4. An example prompt generated after applying the motif encoding modality looks like:  

\begin{tcolorbox}[colback=mycolor,boxsep=1pt,left=4pt,right=4pt,top=4pt,bottom=4pt]
\small
\noindent\textbf{Task}: Node Label Prediction (Predict the label of the node marked with a $?$) given the node-label mapping and graph motif information in the text enclosed in triple backticks. The response should be in the format ``Label of Node = \textless predicted label\textgreater''. If the predicted label cannot be determined, return ``Label of Node = -1''.\\
\textasciigrave \textasciigrave \textasciigrave
\textbf{Node-Label Mapping}: \{1: A, 2: A, 3: ?\}\\
\textbf{Graph-motif information}: No of triangles: 1| Triangles attached to $?$ Node : [1,2,3]| 
\textasciigrave \textasciigrave \textasciigrave 
\end{tcolorbox}
\begin{table*}[ht!]
\centering
\scriptsize
\begin{tabular}{@{}p{0.16\textwidth}p{0.4\textwidth}p{0.4\textwidth}@{}}
\toprule
\textbf{Type of Motif } & \textbf{Motif Encoding} & \textbf{Description of Motif} \\
\midrule
\textbf{Node-Label Mapping} & \textit{Node to Label Mapping : 
Node 1889: Label 4 | ... Node 1893: Label ?|...} 
& Only the node-label mapping is provided (this gives no connectivity information to LLM)\\
\midrule
\textbf{No. of Star Motifs} & Node to Label Mapping : 
Node 1889: Label 4 | ... Node 1893: Label ?| ...
\textit{Graph motif information: Number of star motifs: 0|} & Star motifs signify centralized networks with influential central nodes, where a central node is connected to others that aren't interlinked. We pass the count of the star motifs present in the graph. \\
\midrule
\textbf{No. of Triangle Motifs} & Node to Label Mapping : 
Node 1889: Label 4 | ... Node 1893: Label ?| ...
\textit{Graph motif information: Number of triangle motifs: 6|} & Triangle motifs (triads connecting three nodes) are foundational in social networks, indicating transitive relationships, community structures, and strong social ties. We pass the count of the triads present in the graph. \\
\midrule
\textbf{No. of Triangle Motifs Attached} & Node to Label Mapping : 
Node 1889: Label 4 | ... Node 1893: Label ?|... \textit{Graph motif information: Triangle motifs attached to $?$ node: [1893,2034,1531], [1893,1531,429]|} & We pass the triangle motifs attached to the $?$ label, which gives an idea of the influential triads connected to the $?$ node. \\
\midrule
\textbf{No. of Star Motifs Attached} & Node to Label Mapping : 
Node 1889: Label 4 | ... Node 1893: Label ?|... \textit{Graph motif information: Star motifs connected to $?$ node: []|} & We pass the star motifs attached to the $?$ label, which gives an idea of the influential nodes connected to the $?$ node. \\
\midrule
\textbf{No. of Star and Triangle Motifs } & Node to Label Mapping : 
Node 1889: Label 4 | ... Node 1893: Label ?|... \textit{Graph motif information: Number of star motifs: 0| Number of triangle motifs: 6|} & We pass the count of the triads and star motifs present in the graph, to give the LLM an idea of the graph structure. \\
\midrule
\textbf{Star and Triangle Motifs attached} & Node to Label Mapping : 
Node 1889: Label 4 | ... Node 1893: Label ?|... \textit{Graph motif information: Triangle motifs attached to ? node: [1893,2034,1531], [1893,1531,429]| Star motifs connected to $?$ node: []|}  & We pass the star motifs and triads attached to the $?$ label, which gives an idea of the influential nodes and triads connected to the $?$ node.\\
\midrule
\textbf{No of cliques $?$ Node is part of} & Node to Label Mapping : 
Node 1889: Label 4 | ... Node 1893: Label ?|... \textit{Graph motif information: Number of cliques in graph: 0| $?$ Node is a part of these cliques: []|} & We pass the number of cliques in the network, which gives an idea of its clustered nature. We also pass the cliques the $?$ label is a part of, which gives an idea of the immediate community of the unlabelled node. \\
\midrule
\textbf{No of cliques $?$ Node is attached to } & Node to Label Mapping : 
Node 1889: Label 4 | ... Node 1893: Label ?|...\textit{Graph motif information: $?$ Node is attached to these cliques: []|}  & We pass the cliques the $?$ label is attached to, which gives an idea of the neighboring influential community of the unlabelled node. \\

\bottomrule
\end{tabular}
\caption{Summary of motif information passed as a part of the motif modality encoder with their associated examples and explanations. The \textbf{Aggregate of all changes} setup combines all of the above motif information to give the LLM a local and global view of the graph being passed.
}
\label{tab:motif_representations}
\end{table*}

\subsubsection{Image Modality}
We use GPT-4V \cite{openai_gpt4v} to process graph images to give LLMs a global perspective of graph structural information. An example prompt generated after applying the image encoding modality is shown below.
\begin{tcolorbox}[colback=mycolor,boxsep=1pt,left=4pt,right=4pt,top=4pt,bottom=4pt]
\small
\noindent \textbf{Task}: Node Label Prediction (Predict the label of the red node marked with a ?, given the \textbf{graph structure information in the image}). The response should be in the format "Label of Node = <predicted label>." If the predicted label cannot be determined, return "Label of Node = -1."

\centering
\includegraphics[width=0.35\linewidth, height=2cm]{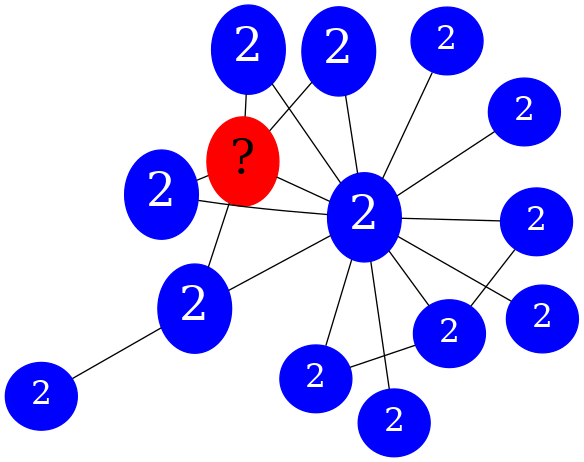} 
\end{tcolorbox}
\begin{table*}[h!]
\centering
\begin{tabular}{p{1.5 cm}|p{1.5cm}|ccc}
\toprule
& \textbf{Model} & \textbf{Cora} & \textbf{Citeseer} & \textbf{Pubmed} \\
\midrule
\multirow{3}{1.5cm}{GNN Baselines} & GCN & 0.7820 \std{0.133} & 0.6540 \std{0.083} & 0.7480 \std{0.077} \\
& GAT & \textbf{0.8200} \std{0.084} & 0.6680 \std{0.069} & 0.7510 \std{0.050} \\
& GraphSage & 0.7570 \std{0.137} & 0.6300 \std{0.098} & 0.7430 \std{0.078} \\
\midrule
\multirow{3}{1.5cm}{LLMs + Encoding Modality} & Text & \underline{0.81} \std{0.04} \small [0.07 \std{0.03}]  & \textbf{0.75} \std{0.05} \small [0.07 \std{0.01}] & \textbf{0.83} \std{0.01} \small [0.08 \std{0.01}]\textsuperscript{*}\\

& Motif & 0.73 \std{0.06} \small [0.06 \std{0.01}] & 0.59 \std{0.01} \small [0.32 \std{0.02}] & 0.77 \std{0.006} \small [0.13 \std{0.04}]\\

& Image & 0.77 \std{0.05} \small [0.04 \std{0.02}]\textsuperscript{*} & \underline{0.71} \std{0.09} \small [0.06 \std{0.0}]\textsuperscript{*} & \underline{0.79} \std{0.03} \small [0.19 \std{0.01}] \\
\bottomrule
\end{tabular}\caption{ Test accuracy rates of node classification across different datasets using the entire 1000 test data and denial rates $D$ in [brackets] for LLM models. For LLMs, we chose a test sample of 50 graphs. \textsuperscript{*} indicates the lowest denial rate for each modality. The highest accuracy rate for the dataset is in bold, while the second highest is underlined.} 
\label{tab:accuracy_tab_old}
\end{table*}

\begin{table}[H]
\centering
\setlength{\tabcolsep}{4pt} 
\begin{tabular}{lccc}
\toprule
\textbf{Details} & \textbf{GCN} & \textbf{GAT} & \textbf{GraphSAGE} \\
\midrule
Epochs & 100 & 100 & 100 \\
Learning Rate & 0.005 & 0.005 & 0.005 \\
Weight Decay & 5e-4 & 5e-4 & 5e-4 \\
\bottomrule
\end{tabular}
\caption{List of GNN training hyperparameters} 
\label{tab:GNN_params}
\end{table}

\section{GNN Experiments}
\label{sec:B}
For training our GNN models, we used the best-found hyperparameters reported in Table \ref{tab:GNN_params}. The training and test set details are given in Table \ref{tab:GNN_traintest_params}. In the main paper, we report the GNN test accuracy on 50 samples due to API constraints by GPT-4, which do not allow us to conduct the apple-to-apple comparison with 1000 test samples. However, Table \ref{tab:accuracy_tab_old} reports the test accuracy for GNN reported on 1000 samples. 
\begin{table}[H]
\centering
\footnotesize 
\setlength{\tabcolsep}{2pt} 
\begin{tabular}{lccc}
\toprule
\textbf{Dataset} & \textbf{Cora} & \textbf{Citeseer} & \textbf{Pubmed} \\
\midrule
Training Set & 140 & 120 & 60 \\
Testing Set & 1000 & 1000 & 1000 \\
GCN Params & 23063 & 59366 & 8067 \\
GAT Params & 92373 & 237586 & 32393 \\
GraphSage Params & 46103 & 118710 & 16115 \\
\bottomrule
\end{tabular}
\caption{GNN Train-Test split and Parameters} 
\label{tab:GNN_traintest_params}
\end{table}